\documentclass[11pt,a4paper]{article}
\usepackage[normalem]{ulem}
\usepackage[dvipsnames]{xcolor}
\usepackage{microtype}
\usepackage{graphicx}
\usepackage{subcaption}
\usepackage{booktabs}
\usepackage{mathtools}
\usepackage{hyperref}
\usepackage{amsfonts}
\usepackage{amsmath}
\usepackage{array, makecell}
\usepackage{multirow}
\usepackage{float}
\usepackage{amssymb}
\usepackage[noend]{algorithmic}
\usepackage{algorithm}

\def\doubleunderline#1{\underline{\underline{#1}}}

\usepackage[hyperref]{emnlp2020}
\usepackage{times}
\usepackage{latexsym}

\aclfinalcopy 
\def\aclpaperid{1376} 

\newcommand{\para}[1]{\smallskip\noindent\textbf{#1}}

\title{Teaching Machine Comprehension with Compositional Explanations}

\author{Qinyuan Ye, Xiao Huang, Elizabeth Boschee, Xiang Ren}

\author{
Qinyuan Ye\textsuperscript{1}~~
Xiao Huang\textsuperscript{1}~~
Elizabeth Boschee\textsuperscript{2}~~
Xiang Ren\textsuperscript{1,2}
\\
\textsuperscript{1}Department of Computer Science, University of Southern California\\
\textsuperscript{2}Information Science Institute, University of Southern California\\
\small{
\texttt{\{qinyuany, huan183\}@usc.edu},~
\texttt{boschee@isi.edu},~
\texttt{xiangren@usc.edu}
}
}

\begin{document}
\captionsetup{font=small}
\maketitle
\begin{abstract}
Advances in machine reading comprehension (MRC) rely heavily on the collection of large scale human-annotated examples in the form of (question, paragraph, answer) triples. In contrast, humans are typically able to generalize with only a few examples, relying on deeper underlying world knowledge, linguistic sophistication, and/or simply superior deductive powers. In this paper, we focus on ``teaching'' machines reading comprehension, using a small number of semi-structured explanations that explicitly inform machines \textit{why} answer spans are correct. 
We extract structured variables and rules from explanations and compose \textit{neural module teachers} that annotate instances for training downstream MRC models. We use learnable neural modules and soft logic to handle linguistic variation and overcome sparse coverage; the modules are jointly optimized with the MRC model to improve final performance.
On the SQuAD dataset, our proposed method achieves 70.14\% F1 score with supervision from 26 explanations, comparable to plain supervised learning using 1,100 labeled instances, yielding a 12x speed up\footnote[1]{\small Our code and data can be found at \url{https://github.com/INK-USC/mrc-explanation}.}.
\end{abstract}

\section{Introduction}

\begin{figure*}
\vspace{-0.1cm}
    \hspace{-0.2cm}
    \includegraphics[width=1.02\textwidth,clip,trim={0cm 12cm 4cm 0}]{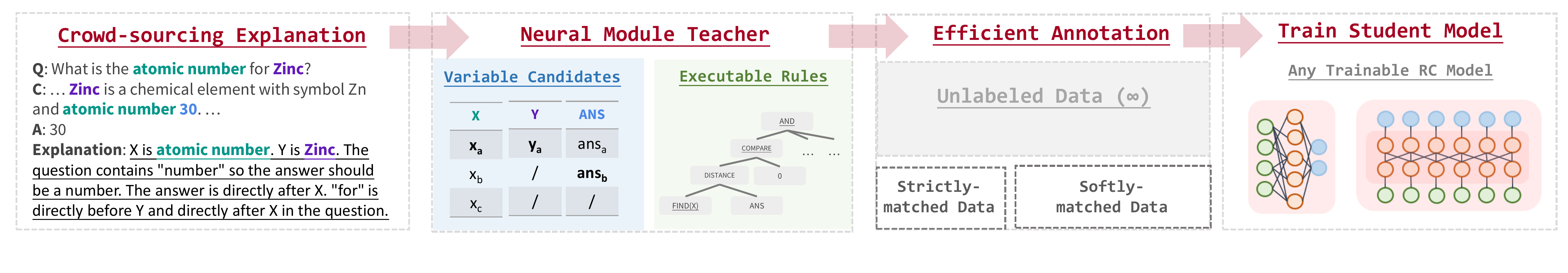}
    \vspace{-1cm}
    \caption{\small\textbf{Overview of proposed work.} We first collect a small set of semi-structured explanations, from which we extract key information such as variables and rules. These structured results are formulated into programs called neural module teachers (NMTeachers), which we use to curate supervision for training machine reading comprehension models.}
    \label{fig:overview}
    \vspace{-0.3cm}
\end{figure*}

\begin{table}[h]
\vspace{-0.2cm}
    \centering
    \scalebox{0.7}{
    \begin{tabular}{p{10cm}}
    \toprule
        \makecell[l{p{10cm}}]{\textbf{\underline{Reference Instance}}\\\textbf{Q:} When was Queen Victoria's \textcolor{Blue}{\uline{\textbf{\textit{funeral}}}} \textcolor{BrickRed}{\uwave{\textbf{held}}}?\\\textbf{C:} Her \textcolor{Blue}{\uline{\textbf{\textit{funeral}}}} was \textcolor{BrickRed}{\uwave{\textbf{held}}} on Saturday, 2 February, in St George's Chapel, Windsor Castle, and after two days of lying-in-state ...\\\textbf{A:} Saturday, 2 February}  \\
        \makecell[l{p{10cm}}]{\textbf{\underline{Semi-structured Explanation}}\\X is ``\textcolor{Blue}{\uline{\textbf{\textit{funeral}}}}''. Y is ``\textcolor{BrickRed}{\uwave{\textbf{held}}}''. In the question X is within 4 words after ``when was'' and Y is directly after X. ``on'' is directly before the answer. Y is within 2 words before the answer. X is \textcolor{PineGreen}{\textbf{within 3 words}} left of Y. The question starts with ``when'', so the answer should be a date.} \\
        \makecell[l{p{10cm}}]{\textbf{\underline{Strictly-matched Instance}}\\\textbf{Q:} When was \textcolor{Blue}{\uline{\textit{\textbf{independence}}}} \textcolor{BrickRed}{\uwave{\textbf{declared}}}?\\\textbf{C:} ... \textcolor{Blue}{\uline{\textit{\textbf{Independence}}}} was \textcolor{BrickRed}{\uwave{\textbf{declared}}} on 24 September 1973.\\\textbf{A:} 24 September 1973} \\
        \makecell[l{p{10cm}}]{\textbf{\underline{Softly-matched Instance}}\\\textbf{Q:} When was \textcolor{Blue}{\uline{\textit{\textbf{Brazelton}}}} \textcolor{BrickRed}{\uwave{\textbf{killed}}}?
        \\\textbf{C:} ... \textcolor{Blue}{\uline{\textit{\textbf{Brazelton}}}} was eventually tracked down and \textcolor{BrickRed}{\uwave{\textbf{killed}}} on Monday August 19, 1878, in a mesquite bosque ...
        \\\textbf{A:} Monday August 19, 1878 (Confidence $z=93.75\%$)}\\
        \textbf{Note:} X is 5 words left of Y, slightly violating ``\textcolor{PineGreen}{\textbf{within 3 words}}''.\\
    \bottomrule
    \end{tabular}
    }
    \vspace{-0.2cm}
    \caption{\small\textbf{Key elements in proposed work.} Semi-structured explanations characterize \textit{why} an answer is correct and summarize the human's deductive process. Strictly and softly matched instances are automatically generated from explanations and provide supervision for training MRC models.}
    \label{tab:example}
    \vspace{-0.4cm}
\end{table}

Recent advances in neural sequence learning and pre-trained language models yield strong (human-level) performance on several reading comprehension datasets~\cite{lan2019albert,raffel2019exploring}. However, state-of-the-art results mainly rely on large-scale annotated corpora, which are often time-consuming and costly to collect~\cite{rajpurkar2016squad}. This often leads to a large gap between methods in the research settings and practical use cases, as large amounts of annotated data rarely exist for a new task or a low-resource domain~\cite{linzen2020can}. To reduce this dependency on annotation efforts, we seek to improve the efficiency in obtaining and applying human supervision. 

One strength of human cognition is the ability to generalize from relatively few examples; shown only a few instances of a problem and solution, humans often deduce patterns more readily than a machine, typically by bringing to bear a wealth of background information about what ``really matters'' in each example~\cite{dejong1986explanation,goldwasser2014learning,lake2019human}. This ability to quickly abstract ``deduction rules'' is the inspiration for this work, and we aim to gather these rules in the form of semi-structured explanations.

In this paper, we focus on the extractive machine reading comprehension (MRC) task, where the system is given a query and is asked to identify an answer span from a particular paragraph. Previous work soliciting explanations as part of the annotation process has been limited to classification tasks \cite{hancock2018training, wang2019learning}. However, MRC is more challenging, since it lacks explicit anchor words (\textit{e.g.}, subject and object in relation extraction), has no pre-defined set of labels, and there is sparser coverage for each explanation.

To tackle these challenges, we propose the concept of a Neural Module Teacher (NMTeacher)~--~an executable \textit{program} constructed from human-provided, semi-structured explanations that is (1) dynamically composed of modules based on the explanation; (2) capable of taking sequential steps and combinatorial search; and (3) capable of fuzzy matching using softened constraints. Fig. \ref{fig:overview} shows an overview of our approach. We first use a Combinatory Categorial Grammar parser \cite{zettlemoyer2012learning} to turn explanations into structured \textit{variables} and \textit{rules} (Sec.~\ref{sec:exp_parse}). A neural module teacher is constructed with basic learnable modules (Sec.~\ref{sec:modules}) based on parsing results and functions as a weak model for the specific type of question described in the explanation (Sec.~\ref{sec:get_answer}). All neural module teachers act together and identify strictly- and softly-matched instances from an unlabeled corpus, which are used to train a downstream ``student'' MRC model (Sec.~\ref{sec:downstream}). It is important to note that while this work is tied to the particular task of MRC, we believe it can be extended to a wide range of NLP tasks. 

We evaluated our approach on two datasets in MRC setting: SQuAD v1.1 \cite{rajpurkar2016squad} and Natural Questions \cite{kwiatkowski2019natural}. Experimental results show the efficiency of the proposed approach in extremely low-resource scenarios. Using 26 explanations gathered in 65 minutes, NMTeacher achieves 56.74\% exact match and 70.14\% F1 score on the SQuAD dataset, while the performance is 9.71\% and 16.37\% with traditional annotation using the same amount of time. Moreover, our analysis shows that explanations continue to improve model performance when a medium-sized annotated dataset is readily available.

\section{Problem Formulation}
\label{sec:formulation}

Our goal is to efficiently train an extractive MRC model $\mathbb{F}$, which takes as input a tuple $(q,c)$ of question $q$ and context $c$, and extracts an answer span $a$ within the context $c$. 
We assume a low-resource situation where a large set $\mathcal{S}$ of $(q,c)$ pairs (without answer annotation) already exists, but we are limited in time to annotate only a small subset $\mathcal{S}_o$ ($< 200$ instances) of $\mathcal{S}$. 

\smallskip
\noindent \textbf{Overview and Notations.} We provide an overview of our proposed method in Fig. \ref{fig:overview}. First, we collect an answer $a_i$ and an explanation $e_i$ for each $(q_i,c_i)$ instance in $\mathcal{S}_o$, resulting in an updated $\mathcal{S}_o=\{(q,c,a,e)\}$. A neural module teacher $\mathbb{G}_i$ will be constructed from each explanation $e_i$, enabling it to answer questions similar to $(q_i, c_i)$. All neural module teachers acting together can be viewed as an ensemble teacher $\mathbb{G}$. We then apply $\mathbb{G}$ to unannotated $(q,c)$ pairs in $\mathcal{S}$, getting $\mathcal{S}_a=\{(q,c,a)\}$, a strictly-labeled dataset that $\mathbb{G}$ can directly answer. The remaining unmatched instances are denoted as $\mathcal{S}_u=\{(q,c)\}$.
After softening the constraints in each $\mathbb{G}_i$, we get a noisily-labeled dataset $\mathcal{S}_p=\{(q,c,a,z)\}$ from $\mathcal{S}_u$, where $z$ is a confidence score given by $\mathbb{G}$. Notably, we will refer to the $(q_i,c_i,a_i)$ part in $(q_i,c_i,a_i,e_i)\in\mathcal{S}_o$ as the ``reference instance'' for explanation $e_i$, since we will frequently check $(q_i,c_i,a_i)$ ``for reference'' when we apply $\mathbb{G}_i$ to new, unseen instances.

$\mathcal{S}_a$ and $\mathcal{S}_p$ are significantly larger in size than $\mathcal{S}_o$ and thus provide more sufficient supervision. We use $\mathcal{S}_a$ and $\mathcal{S}_p$ to train a downstream MRC model $\mathbb{F}$. We denote this method as NMTeacher-DA. We further explore several variants, such as (1) leveraging $\mathcal{S}_u$ with semi-supervised methods; and (2) joint training of $\mathbb{G}$ and $\mathbb{F}$. We construct our final model NMTeacher-Joint by incorporating these variants. Note that our approach is model-agnostic so that $\mathbb{F}$ can take any trainable form. 
\section{Neural Module Teacher}

\label{sec:nmteacher}
A neural module teacher (NMTeacher) acts as a \textit{program} that tries to answer questions following an explanation. In this section, we introduce the basic modules used for rule execution (Sec.~\ref{sec:modules}), discuss how variables and rules are obtained from explanations (Sec.~\ref{sec:exp_parse}), and present how a neural module teacher derives answers (Sec.~\ref{sec:get_answer}).

\subsection{Atomic Modules}

\begin{table}[]
    \centering
    \scalebox{0.7}{
    \begin{tabular}{p{10.5cm}}
    \toprule
        \makecell[l{p{10.5cm}}]{{\large \textsc{Fill} Module}: $(s_{ref}, p_{ref}, s) \rightarrow p$}  \\
        \makecell[l{p{10.5cm}}]{\textbf{Description:} Select the span $p$ in a given sentence $s$ that plays the same syntactic role of span $p_{ref}$ in sentence $s_{ref}$.} \\
        \makecell[l{p{10.5cm}}]{\textbf{Example:} $s_{ref}$ = How is \underline{packet switching} characterized?\\\hspace{1.6cm} $p_{ref}$ = [2,3] (packet switching)\\\hspace{1.6cm} $s$ = How is \underline{hunting} regulated?\\\hspace{1.6cm} $\rightarrow p$ = [2,2] (hunting)} \\\midrule
        \makecell[l{p{10.5cm}}]{{\large \textsc{Find} Module}: $(q_{ref}, p_{ref}, s) \rightarrow p$}  \\
        \makecell[l{p{10.5cm}}]{\textbf{Description:} Find the span $p$ in a context sentence $s$ that refers to the span $p_{ref}$ in the question $q_{ref}$.} \\
        \makecell[l{p{10.5cm}}]{\textbf{Example:} $q_{ref}$ = How is \underline{a promoter sequence} recognized?\\\hspace{1.6cm} $p_{ref}$ = [2,4] (a promoter sequence)\\\hspace{1.6cm} $s$ = The \underline{promoter} is recognized and bound by ...\\\hspace{1.6cm} $\rightarrow p$ = [1,1] (promoter)} \\\midrule
        \makecell[l{p{10.5cm}}]{{\large \textsc{Compare} Module}: $(d_0, d_1) \rightarrow p$}  \\
        \makecell[l{p{10.5cm}}]{\textbf{Description:} Softly evaluate the statement $d_1 \leq d_0$.} \\
        \makecell[l{p{10.5cm}}]{\textbf{Example:} $d_0=0, d_1=1 \rightarrow p=0.75$; \ 
        $d_0=4, d_1=2 \rightarrow p=1$} \\\midrule
        \makecell[l{p{10.5cm}}]{{\large \textsc{LogicAnd} Module}: $(p_1, p_2) \rightarrow p$}  \\
        \makecell[l{p{10.5cm}}]{\textbf{Description:} Perform soft logic \textsc{And} to two scalar probabilities.} \\
        \makecell[l{p{10.5cm}}]{\textbf{Example:} $p_1=0.9, p_2=0.8 \rightarrow p=0.7$; \ $p_1=1, p_2=1 \rightarrow p=1$} \\
    \bottomrule
    \end{tabular}
    }
    \vspace{-0.2cm}
    \caption{\textbf{Summary of atomic modules used in rule execution.} Rules constructed from explanations internally call these modules to fulfill complex functionalities. For example, \textsc{Left}$(X,Y)$ is transformed to \textsc{Compare}(\textsc{Distance}(\textsc{Find}$(X)$, \textsc{Find}$(Y)$), $0$)}
    \label{tab:modules}
    \vspace{-0.5cm}
\end{table}
\label{sec:modules}
We define four types of atomic modules that can be composed to create neural module teachers: \textsc{Fill}, \textsc{Find}, \textsc{Compare} and \textsc{Logic}. Each can support strict and softened matching criteria as a part of generating training instances for downstream use. We summarize their usage in Table~\ref{tab:modules} and introduce them in detail in the following.

\paragraph{\textsc{Fill}.}
When humans encounter a new question, they can detect structural similarities to previous questions. For example, humans will note that \textit{How is \uline{hunting} \uwave{regulated}?} is structually similar to \textit{How is \uline{packet switching} \uwave{characterized}?}, enabling them to infer that answers to both might have a similar structure (\textit{e.g.},\ \textit{by doing sth...}).
To mimic this human intuition, we design a \textsc{Fill} module: given a sentence $s_{ref}$ and a span of interest $p_{ref}$, \textsc{Fill} will predict analogous spans $p$ in a new sentence $s$. 

The strict version of \textsc{Fill} outputs spans $p$ whose named entity type, dependency parse structure, or constituent parse structure\footnote{Identified using spaCy (\url{https://spacy.io/})} matches $p_{ref}$. We encourage over-generation, since the rule execution step later on will verify each candidate. When strict matching produces nothing, we employ softened matching techniques. Here, we first produce a contextualized phrase representation $\mathbf{e}'$ for $p_{ref}$. We rank each candidate constituent $p$ in sentence $s$ according to the similarity between $\mathbf{e}'$ and an analogous phrase representation $\mathbf{e}$ for $p$. We return the top $k$ such constituents along with their score.

To generate phrase representations, we first encode the sentence with BERT-base model \cite{devlin2019bert} and get representations $[\mathbf{h}_1, \mathbf{h}_2, ..., \mathbf{h}_m]$ for each token. We then apply pooling over all tokens in span $p$ to get the phrase representation $\mathbf{e}$. We considered both mean pooling and attentive pooling \cite{bahdanau2014neural}. The similarity score between $\mathbf{e}$ and $\mathbf{e}'$ can be calculated using either cosine similarity or learned bilinear similarity, \textit{i.e.}, $\text{Sim}(\mathbf{e}, \mathbf{e}') = \tanh (\mathbf{e}\mathbf{A}\mathbf{e'} + b)$, where $\mathbf{A}$ is a learnable matrix. We discuss pre-training and design choices for softened \textsc{Fill} module in Sec.~\ref{sec:fill_find_pretrain}.

\paragraph{\textsc{Find}.}
The \textsc{Fill} module finds a span $p$ that plays the same \textit{role} as $p_{ref}$ in its containing sentence. In contrast, \textsc{Find} looks for a span $p$ that has the same \textit{meaning} as $p_{ref}$. For instance, if a query mentions \textit{the explosion}, we might want to identify \textit{exploded} as its counterpart in the paragraph being searched for an answer. This module is similar to the find module in \citet{jiang2019self} in its motivation, while we design ours to be a ranking-based module with discrete boundaries, so that the output fits in the search procedure in Sec. \ref{sec:get_answer}.

The strict version of \textsc{Find} module directly looks for exact matches of the key $p_{ref}$. To account for synonyms, co-reference, and morphological/spelling variation, we also build a softened version using the same model structure as the \textsc{Fill} module. We discuss the design choices and training for the softened \textsc{Find} module in Sec.~\ref{sec:fill_find_pretrain}.

\paragraph{\textsc{Compare}.} In our annotation guidelines, we encourage annotators to describe the relative location of spans in their explanations, \textit{e.g.},\ \textit{X is within 3 words after Y}. The \textsc{Compare} module executes such distance comparisons. The strict version requires the condition to be met exactly: $P(d_1\leq d_0)=1$ when $d_1\leq d_0$, and $P(d_1\leq d_0)=0$ otherwise. In the softened version, we attempt instead to indicate how close $d_1\leq d_0$ is to being true:

{\small
\begin{equation}
P(d_1\leq d_0)=
\begin{cases}
1&d_1 \leq d_0;\\
\max(1-\frac{1}{4}(\frac{d_1-d_0}{|d_0|+1})^2, 0)& d_1 > d_0.
\end{cases}
\label{eq:compare}
\end{equation}
}

As an example, $P(1\leq 0)=0.75$ (a near miss) but $P(5\leq 0)=0$ (due to the $\max$ in Eq. (\ref{eq:compare})).

\paragraph{\textsc{Logic}.} The logic operations ``and'' and ``or'' often appear in explanations. A single explanation may also contain multiple sentences, requiring a logical \textsc{And} to aggregate them. In the strict version of \textsc{Logic}, only boolean outputs of True (1) and False (0) are allowed. In the softened version, we use soft logic to aggregate two probabilities, \textit{i.e.}, $\textsc{And}(p_1,p_2) = \max(p_1+p_2-1,0)$ and $\textsc{Or}(p_1,p_2) = \min(p_1+p_2, 1)$.

\subsection{Parsing Explanations to Executable Rules}
\label{sec:exp_parse}

When soliciting explanations, we encourage annotators to think of each explanation as a collection of \textbf{variables} and \textbf{rules}. This framing allows us to effectively transform these explanations into executable forms. We formally define the terms here:  

\noindent\textbf{Variables} are phrases that may be substituted in a question or answer when generalizing to unseen instances. In Table~\ref{tab:example}, underlined and colored phrases are all considered variables. Annotators are guided to mark these spans explicitly,
\textit{e.g.}, \textit{X is \uline{funeral}. Y is \uline{held}. X is within 5 words of Y}. Variables are closely related to the design of the \textsc{Fill} module since \textsc{Fill} aims to propose potential assignments to these variables when it is given unseen instances.

\noindent\textbf{Rules} are statements that describe the characteristics of variables and relationships between them. When all variables in a rule are assigned, execution of a rule will output either True or False (strict) or a score between 0 and 1 (softened).
Following previous work \cite{srivastava2017joint, wang2019learning}, we first use a Combinatory Categorial Grammar (CCG) based semantic parser $\mathbb{P}$ \cite{zettlemoyer2012learning} to transform explanations into logical forms (\textit{e.g.}, from $e$ to $p_j$ in Table~\ref{tab:parsing}). We build a domain-specific lexicon for common expressions used in explanations. We then implement the operation for each supported predicate (\textit{e.g.}, ``@Is'', ``@Direct'', ``@Left''), which may internally call atomic modules described in Sec \ref{sec:modules}. These predicate implementations, together with the inherent $\lambda$-calculus hierarchy from CCG, will yield the final executable function $f_j$ as shown in Table \ref{tab:parsing}. 

\begin{table}[tb]
    \centering
    \scalebox{0.75}{
    \begin{tabular}{p{9cm}}
    \toprule
        \makecell[l{p{9cm}}]{\textbf{Explanation $e$:} The answer is directly after X.\\\textbf{Parse $p_{j}$:} @Is(Answer, @Direct(@Right(X)))\\\textbf{Execution $f_j$:} \textsc{Compare}(\textsc{Distance}(Ans,\textsc{Find}(X)),0)}\\
        \midrule
        \makecell[l{p{9cm}}]{\textbf{Explanation $e$:} The answer is within 3 words before Z and within 4 words after Y.\\\textbf{Parse $p_{j}$:} @Is(Answer,@And(@LessThan(@Left(Z), 3), \\\quad\quad @LessThan(@Right(Y, 4)))\\\textbf{Execution $f_j$:} \textsc{And}(\textsc{Compare}(\textsc{Distance}(\textsc{Find}(Z),Ans),3),\\\quad\quad\textsc{Compare}(\textsc{Distance}(Ans,\textsc{Find}(Y)),4))}\\
    \bottomrule
    \end{tabular}
    }
    \vspace{-0.2cm}
    \caption{\small\textbf{Rules in three equivalent forms: explanation, parse and underlying execution.} Semi-structured explanations are first parsed and later transformed to executable functions. The execution form is composed of atomic modules (Sec. \ref{sec:modules}). 
    }
    \label{tab:parsing}
    \vspace{-0.4cm}
\end{table}

\subsection{Extracting Answer Spans}
\label{sec:get_answer}

Rules introduced in Sec \ref{sec:exp_parse} can be executed to \textit{verify} whether variable assignments are correct. In other words, given a $(q,c,a)$ triple, executing all rules will give a boolean value (strict) or a confidence score (softened) indicating the triple's correctness.
To actively \textit{output} an answer, we need to re-formulate the problem so that each neural module teacher $\mathbb{G}_i$ takes $(q,c)$ as input and gives an answer span $a$ and confidence score $z$ as output. To this end, we formulate the task of extracting the best answer span into a combinatorial search problem, \textit{i.e.}, searching for the best combination of variable assignments (including the answer).

To apply explanation $e_i$ to a new question, candidates for each variable are first proposed by the \textsc{Fill} module. We then look for the best combination of variable assignments (achieving highest confidence) when evaluated using the rules generated from $e_i$. As a minimal example, if \textsc{Fill} proposes $\{x_1, x_2\}$ as potential assignments to variable X, and $\{a_1, a_2\}$ to ANS, we evaluate the four possible combinations $\{(x_1, a_1), (x_2, a_1) , (x_1, a_2), (x_2, a_2)\}$ by applying $e_i$ and select the one achieving the highest confidence score.
As the number of combinations expands significantly with the number of variables and their candidates, we solve this problem with beam search, progressively filling each variable and in each step keeping the most promising combinations (see Figure \ref{fig:beam} and Algorithm \ref{alg:beam2} in Appendix for more details). By doing so, we have completed our construction of neural module teacher $\mathbb{G}_i$ from one semi-structured explanation $e_i$. We use $\mathbb{G}_i(q,c)=(a,z)$ to denote that given question $q$ and context $c$, neural module teacher $\mathbb{G}_i$ identifies the answer span $a$ with a confidence score $z$. Multiple neural module teachers $\mathbb{G}_i$ may ensemble into $\mathbb{G}$ by listing answer spans outputted by each $\mathbb{G}_i$ and selecting the one with the highest $z$.

{
\renewcommand{\algorithmiccomment}[1]{// #1}
\let\oldReturn\Return
\renewcommand{\algorithmicrequire}{\textbf{Input:}}
\renewcommand{\algorithmicensure}{\textbf{Output:}}
\begin{algorithm}[tb]
\begin{small}
\begin{algorithmic}[1]
\caption{Learning with Explanations}\label{algo:overview}
\REQUIRE{Tiny Dataset $\mathcal{S}_o=\{(q,c)\}$, Large Unlabeled \\Dataset $\mathcal{S}=\{(q,c)\}$, Confidence Threshold $t$}\;
\ENSURE{MRC Model $\mathbb{F}: (q,c)\rightarrow a$}\;
\STATE Collect Ans+Explanation for $\mathcal{S}_o$:  $\mathcal{S}_o\leftarrow\{(q,c,a,e)\}$\;
\STATE\COMMENT{Construct Neural Module Teachers}\;
\STATE $\mathbb{G}\leftarrow \varnothing$\;
\FOR{$(q_i,c_i,a_i)\in \mathcal{S}_o$}
\STATE Parse $e_i$ and construct neural module teacher $\mathbb{G}_i$\;
\IF{$\mathbb{G}_i(q_i,c_i)=(a_i, 1.0)$} 
\STATE $\mathbb{G} = \mathbb{G} \cup \{\mathbb{G}_i\}$ \COMMENT{ $\mathbb{G}_i$ is validated}
\ENDIF
\ENDFOR
\STATE\COMMENT{Generate pseudo labels for $\mathcal{S}$}\;
\STATE $\mathcal{S}_a\leftarrow \varnothing$, $\mathcal{S}_p\leftarrow \varnothing$
\FOR{$(q,c)\in \mathcal{S}$}
\STATE $(a, z) = \mathbb{G}(q, c)$ \COMMENT{$z$ is confidence score}
\IF{$z=1$}
\STATE $\mathcal{S}_a\leftarrow\mathcal{S}_a\cup\{(q,c,a)\}$ \COMMENT{Strict Match}
\ELSE
\STATE $\mathcal{S}_u\leftarrow\mathcal{S}_u\cup\{(q,c)\}$ \COMMENT{Unlabeled}
\IF{$z>t$}
\STATE $\mathcal{S}_p\leftarrow\mathcal{S}_p\cup\{(q,c,a,z)\}$ \COMMENT{Softened Match}
\ENDIF
\ENDIF
\ENDFOR
\STATE\COMMENT{Train Downstream MRC Model $\mathbb{F}$}\;
\STATE$\mathbb{F}\leftarrow $Train$(\mathcal{S}_a, \mathcal{S}_p, \mathcal{S}_u)$\;
\STATE \textbf{return} $\mathbb{F}$
\end{algorithmic}
\end{small}
\end{algorithm}
}

\section{Learning to Augment with NMTeacher}

\subsection{Pre-training the Fill and Find Module}
\label{sec:fill_find_pretrain}

The softened \textsc{Fill} module is pre-trained with pairs of (positive) matches $(q_{ref}, s_{ref}, q, s)$ from strictly-matching results $\mathcal{S}_a$, including 99153 questions and 55202 contexts, divided into 70\% train, 10\% dev and 20\% test datasets. We use random constituents in the sentence as negative training examples.
For the \textsc{Fill} module, we evaluated various model designs described in section~\ref{sec:modules} and choose to use attentive pooling and bilinear similarity.

The softened \textsc{Find} module assesses semantic similarity of phrases. We tried various datasets as proxies for pre-training this ability, including coreference resolution results on SQuAD corpus (produced by Stanford CoreNLP \cite{manning2014stanford}) and paraphrase dataset (PPDB \cite{pavlick-etal-2015-ppdb}). 
We manually evaluated \textsc{Find} module performance with $\mathcal{S}_o$, and we observe that using mean pooling and cosine similarity without any pre-training yields the best performance. We conjecture this may be caused by data bias (the training data not aligning with the purpose of the module).
Therefore, we use untrained BERT-base as our \textsc{Find} module to capture semantic similarities. We leave manual evaluation results in Appendix~\ref{sec:additional_exp}.

\subsection{Training the MRC Model $\mathbb{F}$}
\label{sec:downstream}
Our learning framework (Algorithm \ref{algo:overview}) uses our ensemble neural module teacher $\mathbb{G}$ to answer each $(q,c)$ instance in $\mathcal{S}$, resulting in three splits of data instances: a strictly-matched set $\mathcal{S}_a$, a softly-matched dataset $\mathcal{S}_p$ and an unlabeled set $\mathcal{S}_u$. We use these three sets to jointly learn our downstream MRC model and NMTeacher, as described below.

\smallskip
\noindent\textbf{Learning from Strictly-matched Data $\mathcal{S}_a$.} We start by simply treating $\mathcal{S}_a$ as a labeled dataset, and first train the downstream MRC model $\mathbb{F}$ with traditional supervised learning. We compare different MRC models in our experiments. For simplicity, we denote $\texttt{MRC\_Loss}(\mathcal{B}^{(i)})$ as the loss term defined in these MRC models for the $i$-th instance in batch $\mathcal{B}$. In each step, we sample a batch $\mathcal{B}_a$ from $\mathcal{S}_a$ and update the model with loss term $\mathcal{L}(\mathcal{B}_a)$:
\vspace{-0.2cm}

{
\small
\begin{equation}
    \mathcal{L}(\mathcal{B}_a) = \sum_{i=1}^{|\mathcal{B}_a|} \frac{1}{|\mathcal{B}_a|}\cdot \texttt{MRC\_Loss}(\mathcal{B}_a^{(i)}).
\end{equation}
}
\vspace{-0.3cm}

\noindent\textbf{Learning from Softly-matched Data $\mathcal{S}_p$.} The softly-matched set $\mathcal{S}_p$ is significantly larger in size (than $\mathcal{S}_a$) and may contain useful information for training $\mathbb{F}$. We blend in supervision from $\mathcal{S}_p$ by adding a weighted loss term to the original loss $\mathcal{L}(\mathcal{B}_a)$. That is, we simultaneously sample a batch $\mathcal{B}_a$ from $\mathcal{S}_a$ and a batch $\mathcal{B}_p$ from $\mathcal{S}_p$. The loss term for $\mathcal{B}_p$ is weighted and normalized by the confidence score $z$ from NMTeacher $\mathbb{G}$,
\vspace{-0.4cm}

{\small
\begin{align}
    w_i &= \frac{\exp(\theta_t z_i)}{\sum_{j=1}^{|\mathcal{B}_p|}\exp(\theta_t z_j)},\label{eq:norm}\\
\mathcal{L}(\mathcal{B}_p) &= \sum_{i=1}^{|\mathcal{B}_p|} w_i\cdot \texttt{MRC\_Loss}(\mathcal{B}_p^{(i)}),
\end{align}
}

\vspace{-0.4cm}
\noindent where $\theta_t$ in Eq. \ref{eq:norm} is a temperature that controls the normalization intensity. We then aggregate the loss terms from $\mathcal{S}_p$ and $\mathcal{S}_a$ with coefficient $\beta$, \textit{i.e.}, $\mathcal{L}_{ap} = \mathcal{L}(\mathcal{B}_a) + \beta \mathcal{L}(\mathcal{B}_p)$. We denote the method up to this step as NMTeacher-DA.

\smallskip
\noindent\textbf{Learning from Unlabeled Data $\mathcal{S}_u$.} We further learn from unlabeled data in $\mathcal{S}_u$ by integrating existing semi-supervised methods. In brief, pseudo labeling (PL) samples a batch $\mathcal{B}_u$ from $\mathcal{S}_u$, annotates it with the current MRC model $\mathbb{F}$, and calculates the loss term on this pseudo-labeled batch $\mathcal{B}_u$. The overall loss $\mathcal{L}$ term thus becomes $\mathcal{L}_{au} = \mathcal{L}(\mathcal{B}_a) + \beta \mathcal{L}(\mathcal{B}_u)$. To mix in supervision from unlabeled data, we formulate a $r+1$ rotation between sampling unlabeled batch $\mathcal{B}_u$ and softly-matched batch $\mathcal{B}_p$; we update MRC model $\mathbb{F}$ for $r$ steps using the semi-supervised method and loss term $\mathcal{L}_{au}$, and then update the model for one step using softly-matched data and the loss term $\mathcal{L}_{ap}$.

\smallskip
\noindent\textbf{Joint Training.} Instance weight $w_i$ (Eq. \ref{eq:norm}) for each softly-labeled instance in batch $\mathcal{B}_p$ is calculated with NMTeacher $\mathbb{G}$, so we further allow gradient backpropagation to trainable \textsc{Fill} and \textsc{Find} modules in $\mathbb{G}$ when optimizing loss term $\mathcal{L}_{au}$. We fix $\mathbb{G}$ at first and allow joint training after training on $\mathbb{F}$ converges. This helps form consensus between NMTeacher $\mathbb{G}$ and the learned downstream MRC model $\mathbb{F}$, which we believe is helpful in denoising and refining the final MRC model. We denote this final method as NMTeacher-Joint.

\begin{table*}[tb!]
\vspace{-0.2cm}
\centering
\scalebox{0.65}{
\begin{tabular}{@{}lcccccc@{}}
\cmidrule[1pt](){1-7}
\multirow{2}{*}{\textbf{\#Explanations ($|\mathcal{S}_a|$, $|\mathcal{S}_p|$)}} & \multicolumn{2}{c}{13 (131, 314)} & \multicolumn{2}{c}{26 (424, 1048)} & \multicolumn{2}{c}{52 (766, 2329)} \\ \cmidrule(lr){2-3} \cmidrule(lr){4-5} \cmidrule(lr){6-7}
& EM & F1 & EM & F1 & EM & F1 \\ 
\cmidrule{1-7}
BiDAF ($\mathcal{S}_a$)  &   3.66 $\pm$ 0.92 & 7.80 $\pm$ 0.84 & 5.49 $\pm$ 0.50 & 9.91 $\pm$ 0.34 & 8.21 $\pm$ 0.25 & 14.15 $\pm$ 0.40 \\
~~+ NMTeacher-DA ($\mathcal{S}_p$)      & 5.15 $\pm$ 0.45 & 8.51 $\pm$ 0.22 & 6.65 $\pm$ 0.34 & 11.46 $\pm$ 0.49 & 12.63 $\pm$ 0.86 & 19.99 $\pm$ 1.06 \\
\cmidrule{1-7}
BERT-base ($\mathcal{S}_a$)       & 10.52 $\pm$ 1.57 & 17.88 $\pm$ 2.09 & 19.90 $\pm$ 1.53 & 30.42 $\pm$ 1.53 & 28.84 $\pm$ 1.69 & 39.26 $\pm$ 2.12  \\
~~+ NMTeacher-DA ($\mathcal{S}_p$)  & 13.80 $\pm$ 1.29 & 23.39 $\pm$ 1.43 & 22.30 $\pm$ 2.78 & 32.96 $\pm$ 5.00 & 30.74 $\pm$ 2.48 & 41.28 $\pm$ 3.14 \\
\cmidrule{1-7}
BERT-large ($\mathcal{S}_a$)       & 13.27 $\pm$ 1.09 & 21.11 $\pm$ 2.26 & 25.90 $\pm$ 2.55 & 38.35 $\pm$ 2.38 & 34.66 $\pm$ 0.65 & 47.32 $\pm$ 0.60 \\
~~+ NMTeacher-DA ($\mathcal{S}_p$)  & 15.80 $\pm$ 1.64 & 27.45 $\pm$ 2.32 &  28.07 $\pm$ 2.27 & 41.95 $\pm$ 2.95 & 39.05 $\pm$ 1.36 & 51.65 $\pm$ 2.08 \\
~~+ Self Training ($\mathcal{S}_u$)      & 15.25 $\pm$ 2.49 & 23.13 $\pm$ 2.84 & 30.43 $\pm$ 6.30 & 40.80 $\pm$ 4.53 & 43.55 $\pm$ 3.39 & 54.62 $\pm$ 4.40 \\
~~+ Mean Teacher ($\mathcal{S}_u$)      & 11.84 $\pm$ 2.36 & 19.62 $\pm$ 2.37 & 32.80 $\pm$ 5.72 & 45.50 $\pm$ 4.61 & 41.86 $\pm$ 7.22 & 54.74 $\pm$ 5.80 \\
~~+ Pseudo Labeling ($\mathcal{S}_u$)      & 14.82 $\pm$ 1.70 & 21.67 $\pm$ 2.96 & 38.10 $\pm$ 5.62 & 50.62 $\pm$ 7.30 & 50.45 $\pm$ 2.11 & 61.82 $\pm$ 1.32 \\
~~+ NMTeacher-Joint ($\mathcal{S}_p+\mathcal{S}_u$)& 34.80 $\pm$ 14.16 & 44.00 $\pm$ 17.74 & \textbf{56.74 $\pm$ 1.27} & 70.14 $\pm$ 2.58 & \textbf{58.11 $\pm$ 0.95} & 70.67 $\pm$ 1.58 \\
\cmidrule{1-7}
ALBERT-base ($\mathcal{S}_a$)   & 30.12 $\pm$ 1.00 & 42.95 $\pm$ 1.65 & 39.24 $\pm$ 1.80 & 53.40 $\pm$ 2.87 & 44.57 $\pm$ 1.90 & 58.09 $\pm$ 0.59  \\
~~+ NMTeacher-DA ($\mathcal{S}_p$) & 34.31 $\pm$ 1.23 & 46.59 $\pm$ 1.16 & 40.79 $\pm$ 0.78 & 55.22 $\pm$ 0.29 & 46.55 $\pm$ 1.04 & 59.80 $\pm$ 0.64 \\
~~+ Self Training ($\mathcal{S}_u$)  & 35.45 $\pm$ 3.58 & 45.27 $\pm$ 3.71 & 46.21 $\pm$ 3.46 & 58.20 $\pm$ 4.04 & 47.08 $\pm$ 3.70 & 60.57 $\pm$ 4.11 \\
~~+ Mean Teacher ($\mathcal{S}_u$)   & 29.35 $\pm$ 1.79 & 41.73 $\pm$ 1.07 &  40.92 $\pm$ 2.05 & 55.17 $\pm$ 2.36 & 52.16 $\pm$ 0.66 & 65.83 $\pm$ 1.52 \\
~~+ Pseudo Labeling ($\mathcal{S}_u$)   & 27.35 $\pm$ 2.66 & 39.95 $\pm$ 4.24 & 38.56 $\pm$ 2.81 & 51.77 $\pm$ 2.53 & 43.76 $\pm$ 1.88 & 56.69 $\pm$ 2.50 \\
~~+ NMTeacher-Joint ($\mathcal{S}_p+\mathcal{S}_u$) & \textbf{40.67 $\pm$ 5.48} & \textbf{52.49 $\pm$ 4.74} & 54.88 $\pm$ 3.16 & \textbf{70.21 $\pm$ 3.21} & 57.69 $\pm$ 0.77 & \textbf{71.75 $\pm$ 0.48} \\
\cmidrule[1pt](){1-7}
\end{tabular}
}
\vspace{-0.2cm}
\caption{\small \textbf{Performance comparison on SQuAD using 13/26/52 explanations.} $\mathcal{S}_a$ is the set of strictly matched instances annotated by NMTeacher. $\mathcal{S}_p$ is the set of softly matched instances by using softened modules in rule execution. $\mathcal{S}_p$ constantly brings improvements over model trained solely on $\mathcal{S}_a$, showing that the usage of softly-matched but noisy data are beneficial. Such improvement is most significant in extreme low-resource cases with 13 explanations. Best performance is achieved when semi-supervised learning on unlabeled data $\mathcal{S}_u$ and joint training of NMTeacher and MRC model are enabled (NMTeacher-Joint). }
\label{tab:squad-result}
\vspace{-0.1cm}
\end{table*}

\begin{table}[t]
    \centering
    \scalebox{0.73}{
    \begin{tabular}{l|c|c}
        \toprule
        \multicolumn{1}{l|}{\textbf{Statistics / Dataset}} & \multicolumn{1}{c|}{\textbf{SQuAD}} & \multicolumn{1}{c}{\textbf{NQ}} \\
        \midrule
        \# Collected raw explanations & 2,065 & 1,220 \\
        \# Accepted explanations  & 570 & 343 \\
        \# Parsable explanations & 163 & 109 \\
        \# Validated explanations & 130 & 89 \\
        Average \# sentences per explanation & 4.31 & 4.51 \\
        Average \# tokens per explanation & 38.87 & 41.28 \\
        Average \# variables per explanation & 1.96 & 1.47 \\
        \bottomrule
    \end{tabular}
    }
    \vspace{-0.2cm}
    \caption{\small\textbf{Statistics of the collected explanations.}}
    \label{tab:exp_stats}
    \vspace{-0.2cm}
\end{table}

\begin{table*}[t]
\centering
\scalebox{0.65}{
\begin{tabular}{@{}lcccccc@{}}
\cmidrule[1pt](){1-7}
\multirow{2}{*}{\#Explanations ($|\mathcal{S}_a|$, $|\mathcal{S}_p|$)} & \multicolumn{2}{c}{18 (98, 539)} & \multicolumn{2}{c}{36 (107, 647)} & \multicolumn{2}{c}{54 (273, 1047)} \\ \cmidrule(lr){2-3} \cmidrule(lr){4-5} \cmidrule(lr){6-7}
& EM & F1 & EM & F1 & EM & F1 \\
\cmidrule{1-7}
BERT-l ($\mathcal{S}_a$)       & 11.63 $\pm$ 1.52 & 20.86 $\pm$ 1.78 & 15.26 $\pm$ 0.55 & 24.89 $\pm$ 1.47 & 14.24 $\pm$ 0.74 & 24.85 $\pm$ 1.77 \\
~~+ NMTeacher-DA ($\mathcal{S}_p$)  & 17.47 $\pm$ 0.76 & 28.30 $\pm$ 0.42 & 20.77 $\pm$ 2.04 & 31.86 $\pm$ 2.37 & 19.33 $\pm$ 2.44 & 31.56 $\pm$ 2.55 \\
~~+ Self Training ($\mathcal{S}_u$)  & 15.92 $\pm$ 2.13 & 25.17 $\pm$ 0.65 & 18.42 $\pm$ 0.67 & 27.85 $\pm$ 0.46 & 17.49 $\pm$ 1.67 & 26.18 $\pm$ 0.55 \\
~~+ Mean Teacher ($\mathcal{S}_u$)   & 14.67 $\pm$ 0.32 & 24.63 $\pm$ 0.57 & 17.94 $\pm$ 0.93 & 27.71 $\pm$ 0.98 & 17.63 $\pm$ 1.32 & 27.12 $\pm$ 1.24 \\
~~+ Pseudo Labeling ($\mathcal{S}_u$)   & 17.86 $\pm$ 1.71 & 25.47 $\pm$ 0.36 & 20.18 $\pm$ 2.35 & 27.60 $\pm$ 2.40 & 16.56 $\pm$ 0.41 & 25.80 $\pm$ 0.66 \\
~~+ NMTeacher-Joint ($\mathcal{S}_p+\mathcal{S}_u$) & 17.36 $\pm$ 0.70 & 28.36 $\pm$ 1.09 & 23.22 $\pm$ 1.74 & 33.93 $\pm$ 2.16 & 24.04 $\pm$ 2.90 & 34.90 $\pm$ 2.65 \\
\cmidrule{1-7}
ALBERT-b ($\mathcal{S}_a$)   & 19.62 $\pm$ 2.39 & 27.84 $\pm$ 2.89 & 21.78 $\pm$ 2.93 & 31.20 $\pm$ 3.46 & 21.19 $\pm$ 1.80 & 32.08 $\pm$ 1.48 \\
~~+ NMTeacher-DA ($\mathcal{S}_p$) & 21.17 $\pm$ 1.48 & 30.67 $\pm$ 2.47 & 25.93 $\pm$ 3.91 & 35.82 $\pm$ 3.73 & 23.16 $\pm$ 4.26 & 33.89 $\pm$ 3.59 \\
~~+ Self Training ($\mathcal{S}_u$)  & 19.41 $\pm$ 1.31 & 28.04 $\pm$ 1.71 & 22.15 $\pm$ 2.50 & 31.09 $\pm$ 2.30 & 21.65 $\pm$ 2.92 & 31.08 $\pm$ 2.93 \\
~~+ Mean Teacher ($\mathcal{S}_u$)   & 20.26 $\pm$ 0.65 & 29.25 $\pm$ 0.14 & 24.71 $\pm$ 3.38 & 33.66 $\pm$ 3.65 & 28.06 $\pm$ 2.48 & 37.91 $\pm$ 2.15 \\
~~+ Pseudo Labeling ($\mathcal{S}_u$)   & 18.88 $\pm$ 1.98 & 27.28 $\pm$ 1.88 & 23.30 $\pm$ 2.67 & 31.96 $\pm$ 1.46 & 20.23 $\pm$ 1.43 & 30.62 $\pm$ 2.63 \\
~~+ NMTeacher-Joint ($\mathcal{S}_p+\mathcal{S}_u$) & \textbf{24.12 $\pm$ 4.12} & \textbf{34.65 $\pm$ 5.03} & \textbf{30.56 $\pm$ 2.42} & \textbf{41.14 $\pm$ 3.10} &\textbf{ 29.45 $\pm$ 3.64} & \textbf{41.14 $\pm$ 3.14} \\
\cmidrule[1pt](){1-7}
\end{tabular}
}
\vspace{-0.3cm}
\caption{\small \textbf{Performance comparison on NQ using 18/36/54 explanations.} Similar trends as in Table~\ref{tab:squad-result} can be observed.}
\label{tab:nq-result}
\end{table*}

\section{Experiments}

\subsection{Experiment Setup}
\label{sec:exp_setup}
\paragraph{Datasets.} (1) \textbf{SQuAD v1.1} \cite{rajpurkar2016squad} contains over 10k crowd-sourced MRC instances. All questions are answerable. (2) \textbf{Natural Questions (NQ)} \cite{kwiatkowski2019natural} contains questions from Google search queries, paired with related Wikipedia articles. To keep consistent with our settings, we assume ``the long answer is given, and a short answer is known to exist'' and preprocess NQ into the same format as SQuAD. We discard instances whose (1) long answer is not free-form text (\textit{e.g.}, table, list); or (2) short answer contains multiple short spans.

\paragraph{Evaluation.}
Use of the official SQuAD and NQ test sets is restricted, so we construct our own dev and test sets by splitting the official dev sets in half.\footnote{SQuAD: 5537 dev / 5033 test. NQ: 1252 dev / 1252 test.} Hyper-parameters and the best checkpoint are selected on the dev set. We use the  SQuAD official evaluation script and report Exact Match (EM) and F1 score on both the dev set (in Appendix) and test set (in Sec \ref{sec:main_result}). Note that this is different from the long-/short-answer metrics for NQ official evaluation. We report 3-run average and standard deviation using 3 different random seeds.

\paragraph{MRC Models.}
Importantly, our approach is model-agnostic. We test our framework using the following three models as MRC model $\mathbb{F}$. (1) \textbf{BiDAF} \cite{seo2016bidirectional}, which adopts hierarchical architecture and attention mechanism to model question-context interactions; (2) \textbf{BERT} \cite{devlin2019bert}, a pre-trained language model with an additional output layer for MRC\footnote{We use BERT-l as a short hand for BERT-large and BERT-b for BERT-base in following analysis.}; and (3) \textbf{ALBERT} \cite{lan2019albert}, a lite and top-performing model on SQuAD leaderboard. 


\paragraph{Semi-supervised Methods.}
We compare and enhance NMTeacher with the following semi-supervised methods: (1) \textbf{Self Training (ST)} \cite{rosenberg2005semi} iteratively annotates unlabeled instances with maximal confidence in each epoch; (2) \textbf{Pseudo Labeling (PL)} \cite{lee2013pseudo} trains a weak model on labeled data first and annotates unlabeled batches as supervision. (3) \textbf{Mean Teacher (MT)} \cite{tarvainen2017mean} introduces consistency loss between a student model and a teacher model (the exponential moving average of student models from previous steps).

\paragraph{Explanation Collection.}
Table~\ref{tab:exp_stats} provides statistics on the explanations we collected for this effort. We refer readers to Appendix \ref{sec:exp_collect_detail} for more details, including our crowd-sourcing interface and guidelines. 
On average, annotators spend 43 seconds to annotate an answer and 151 seconds to annotate both an explanation and an answer (3.5x slower compared to annotating answer only).

\subsection{Performance Comparison}
\label{sec:main_result}

\begin{figure}[t]
    \centering
    \includegraphics[width=0.38\textwidth,clip,trim={0cm, 0cm, 0cm, 0cm}]{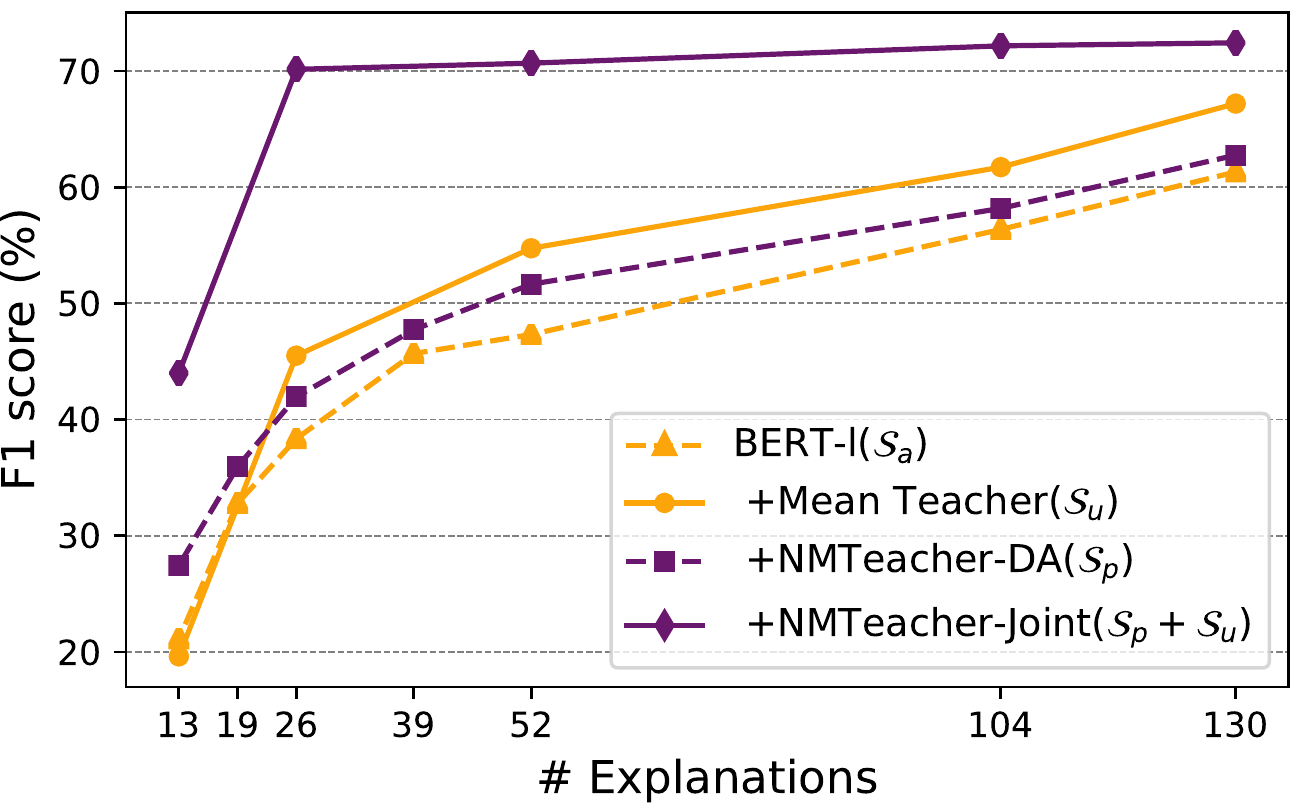}
    \vspace{-0.2cm}
    \caption{\small \textbf{Performance changes with respect to number of explanations on SQuAD.} Performance of the proposed method grow progressively with more explanations.}
    \label{fig:number}
    \vspace{-0.3cm}
\end{figure}

\noindent
\textbf{Main Results.}
Tables \ref{tab:squad-result} and \ref{tab:nq-result} show results of different MRC models, with different numbers of explanations used. The baseline for each model uses as training the strictly-matched instances ($\mathcal{S}_a$) generated using the explanations. For all models, performance then improves when we include the softly-matched instances ($\mathcal{S}_p$). We show in Fig.~\ref{fig:number} that this pattern largely continues even as we further increase the number of explanations, showing that noisy labels are of highest value in low-resource settings but still continue to provide value as training sizes increase. In most cases, performance improves further when trained with semi-supervised learning and $\mathcal{S}_u$. Finally, performance is best when we make full use of $\mathcal{S}_a$, $\mathcal{S}_p$ and $\mathcal{S}_u$, and jointly train $\mathbb{F}$ and $\mathbb{G}$ (NMTeacher-Joint). 

\begin{figure}[t]
    \centering
    \includegraphics[width=0.46\textwidth,clip,trim={0cm, 0cm, 0cm, 0cm}]{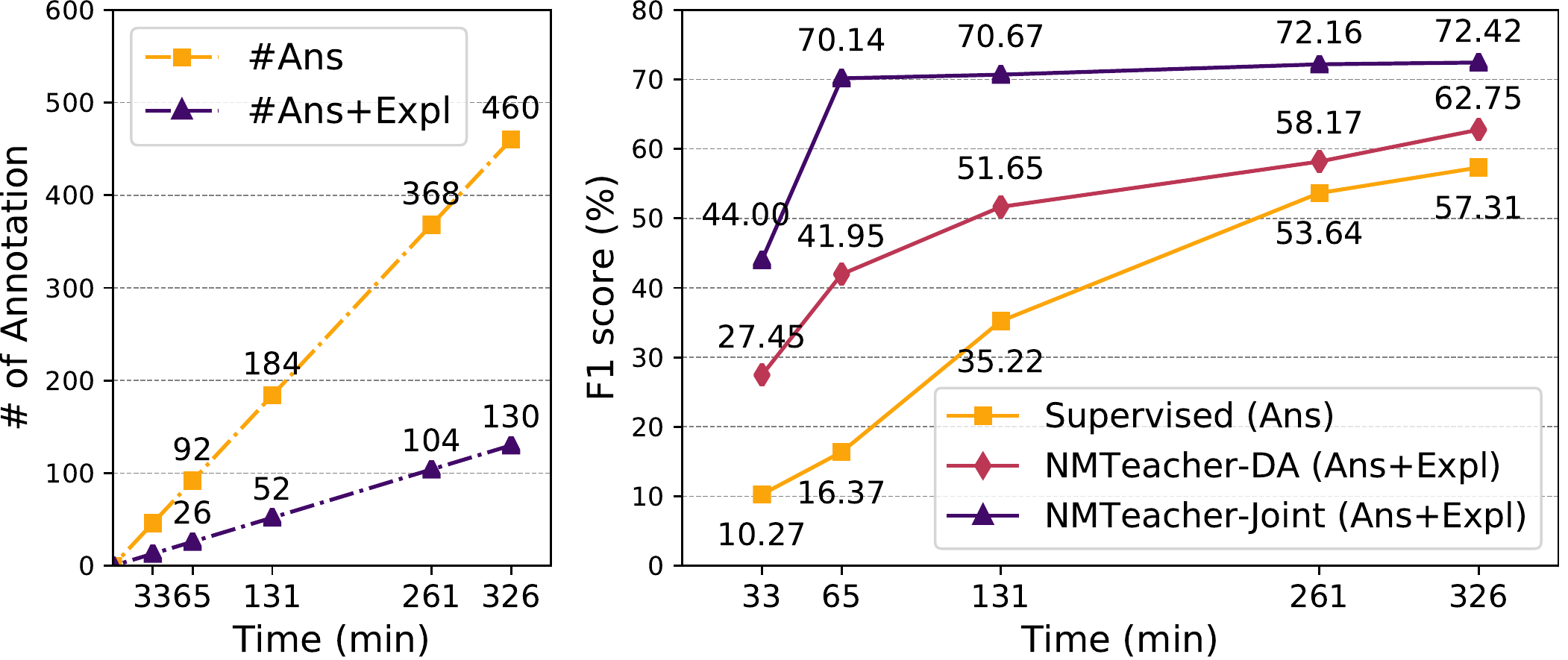}
    \vspace{-0.2cm}
    \caption{\small \textbf{Study on Annotation Efficiency.} We compare model performance when annotation time is held constant;  NMTeacher-Joint consistently outperforms the baseline without explanations (\textit{e.g.},\ 70.14\% vs. 16.37\% F1 score with 65 minutes annotation). BERT-l is used as MRC model.}
    \label{fig:efficiency}
\end{figure}

\para{Efficiency Study.} 
We demonstrate NMTeacher's efficiency by controlling annotation time. 
Given a fixed amount of time $t$, we denote $\mathcal{S}_{r}^{(t)}$ as plain answers that could be collected in $t$;  $\mathcal{S}_a^{(t)}$ and $\mathcal{S}_p^{(t)}$ as strictly and softly matched data generated by answers + explanations collected in $t$. We train a BERT-l MRC model in the following settings: (1) Supervised learning with $\mathcal{S}_r^{(t)}$; (2) NMTeacher-DA with $\mathcal{S}_a^{(t)}$ and $\mathcal{S}_p^{(t)}$; (3) NMTeacher-Joint. Fig.~\ref{fig:efficiency} shows that NMTeacher significantly improves performance over the baseline when annotation time is constant.
Additionally, we found that the 70.14\% F1 score achieved with 26 explanations, requires 1,100 annotated examples if put in supervised learning setting. This gives a 12x annotation speed up.

\begin{table}[t]
\centering
\scalebox{0.65}{
\begin{tabular}{llcc}
\toprule
No. & Training Supervision & EM & F1 \\
\midrule
(1)  & $\mathcal{S}_a$            & 44.57 $\pm$ 1.90   & 58.09 $\pm$ 0.59   \\
(2)  & $\mathcal{S}_a$+$\mathcal{S}_p$            &  46.55 $\pm$ 1.90  & 59.80 $\pm$ 0.64\\
(3)  & $\mathcal{S}_a^*$            &  52.14 $\pm$ 2.02  & 64.25 $\pm$ 1.89    \\
(4)  & $\mathcal{S}_a^*$+$\mathcal{S}_p^*$            & 59.67 $\pm$ 0.33   & 71.55 $\pm$ 0.34    \\
(5)  & $\mathcal{S}_{r} (|\mathcal{S}_{r}|=|\mathcal{S}_a|)$            & 59.15 $\pm$ 0.88   & 71.40 $\pm$ 0.61   \\
(6)  & $\mathcal{S}_{r} (|\mathcal{S}_{r}|=|\mathcal{S}_a|+|{S}_p|)$            &  69.27 $\pm$ 0.30  & 80.09 $\pm$ 0.66  \\
\bottomrule
\end{tabular}}
\vspace{-0.1cm}
\caption{\small\textbf{Analysis on Matching Quality.} $\mathcal{S}_a$ and $\mathcal{S}_p$ are obtained with 52 explanations. $\mathcal{S}_a^*$ denotes instances in $\mathcal{S}_a$ paired with human annotations. $\mathcal{S}_r$ is randomly sampled from SQuAD with size controlled to be equal to $|\mathcal{S}_a|$ or $|\mathcal{S}_a|+|\mathcal{S}_p|$.} 
\label{tab:quality}
\end{table}

\subsection{Performance Analysis}
\label{sec:performance}

\begin{figure}[tb]
\vspace{-0.2cm}
    \centering
    \includegraphics[width=0.45\textwidth,clip,trim={0cm, 0cm, 0cm, 0cm}]{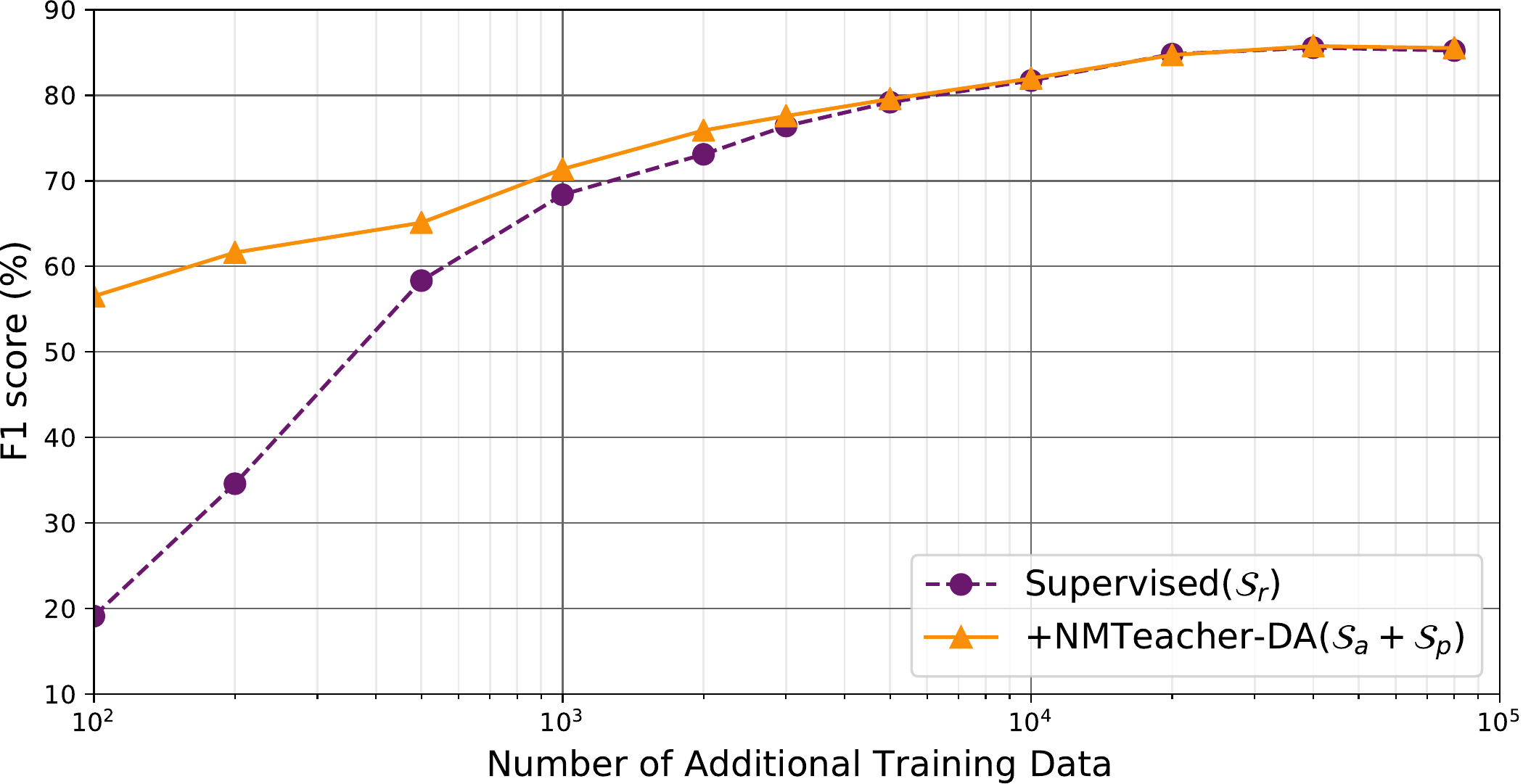}
    \vspace{-0.2cm}
    \caption{\small \textbf{Augmenting Labeled Instances with Explanations in medium-/high-resource scenarios}. Please refer to Sec~\ref{sec:performance} for in-depth analysis.}
    \label{fig:medium_resource}
    \vspace{-0.4cm}
\end{figure}

\noindent
\textbf{Matching Noise/Bias.} 
\label{sec:bias}
Our proposed method hypothesizes new training examples, which may be noisy even when ``strictly matched''. The matched instances may also be more similar than desired to the reference instances. To assess the impact of these two factors, we look at the strictly-matched set $\mathcal{S}_a$ and the softly-matched set $\mathcal{S}_p$ generated with 52 SQuAD explanations. We define $\mathcal{S}_a^*$ and $\mathcal{S}_p^*$, versions of these sets with human-annotated answers (\textit{i.e.}, no noise). We then train an ALBERT-b model with supervision in the following six settings: (1) $\mathcal{S}_a$; (2) $\mathcal{S}_a$ and $\mathcal{S}_p$; (3) $\mathcal{S}_a^*$; (4) $\mathcal{S}_a^*$ and $\mathcal{S}_p^*$; (5) $\mathcal{S}_r$, a set of randomly sampled SQuAD training instances with size $|\mathcal{S}_a|$; (6) $\mathcal{S}_r$ of size $|\mathcal{S}_a|+|\mathcal{S}_p|$. Results are listed in Table \ref{tab:quality}. Comparing (1) and (3), we observe a 6.16\% F1 gap caused by noise in strict matching; Comparing (2) and (4), we see that the gap is further widened, since there are more noises in softly-matched data. Comparing (3) and (5), we see a 7.15\% F1 gap mainly caused by bias in the instances matched by NMTeachers. We believe addressing these two issues will improve model performance, and we leave this as future work.


\para{Medium and High Resource Scenarios.}
Going beyond low-resource scenarios, we examine NMTeacher's capability in medium- and high-resource scenarios. Similar to the few-shot evaluation in \citet{lewis-etal-2019-unsupervised}, we randomly sample different number of human-annotated instances from SQuAD as $\mathcal{S}_r$. The size of $\mathcal{S}_r$ range from 100 to 80k. We train a BERT-l MRC model using $\mathcal{S}_r$ along with $\mathcal{S}_a$, $\mathcal{S}_p$ generated with 52 explanations. We compare with training the MRC model using $\mathcal{S}_r$ only. Fig. \ref{fig:medium_resource} shows that when a medium-size $\mathcal{S}_r$ is readily available ($|\mathcal{S}_r| < 5k$), augmenting it with NMTeacher is still beneficial. In practice, this could be particularly useful when a defect is observed in the trained model (\textit{e.g.}, a certain type of question is answered poorly). A small set of explanations could be collected rapidly and used by NMTeacher to remedy the defect. Benefits brought by NMTeacher become marginal when labeled data set become larger ($|\mathcal{S}_r| > 10k$).

\begin{figure}[t]
    \centering
    \includegraphics[width=0.45\textwidth,clip,trim={0cm, 0cm, 0cm, 0cm}]{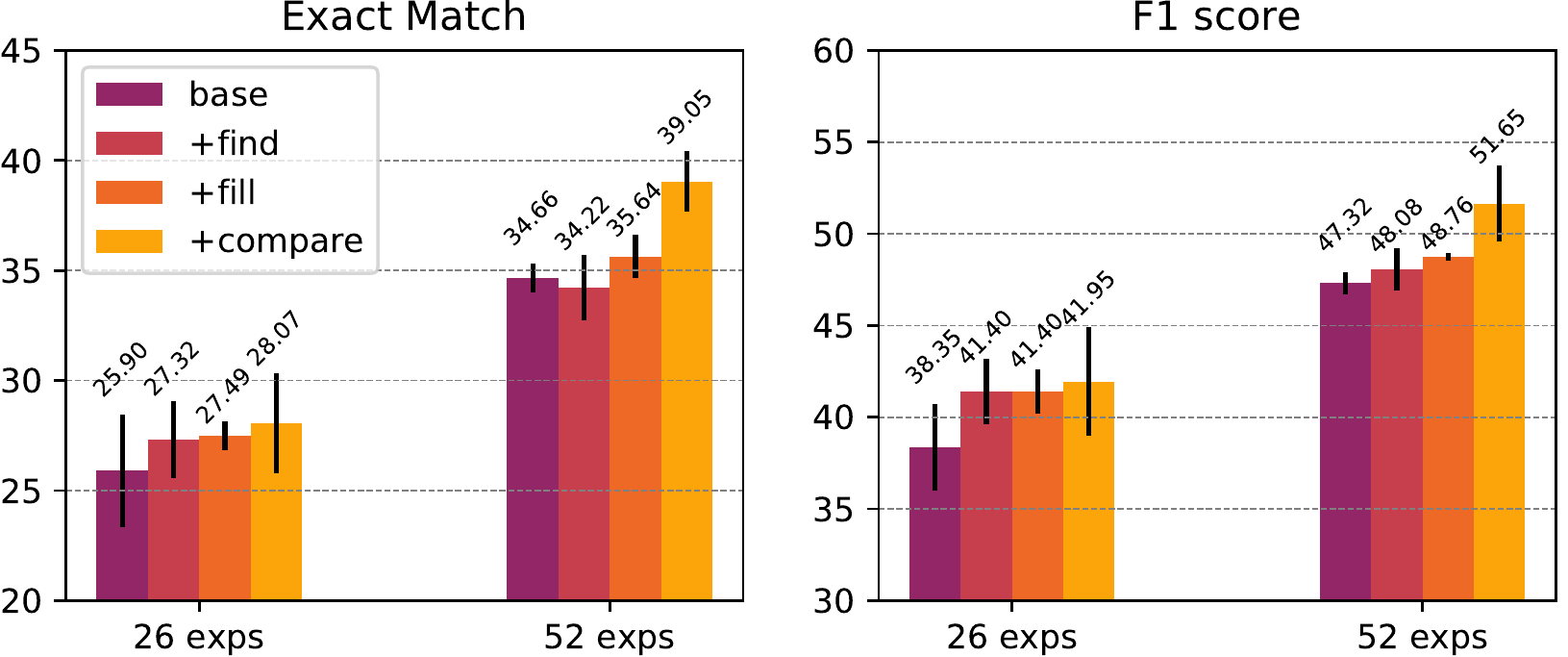}
    \vspace{-0.1cm}
    \caption{\textbf{Ablation study on atomic modules.} Fill, Find and compare modules are switched to softened mode consecutively. Rule softening in each module contributes to improve final MRC model performance.}
    \label{fig:ablation}
    \vspace{-0.4cm}
\end{figure}

\para{Ablation Study on Modules.}
To evaluate the effect of the softened module execution, we progressively turn on the softened version of \textsc{Find}, \textsc{Fill} and \textsc{Compare} in NMTeacher matching process, use matched data to train the downstream MRC model $\mathbb{F}$ in NMTeacher-DA setting, and report the final performance. The evaluation results are presented in Fig.~\ref{fig:ablation}. Results show that softening each module contributes to performance improvement.

\para{Additional Analysis.} We refer readers to Appendix \ref{sec:additional_exp} for additional matching quality analysis, and manual evaluation of trainable modules.

\subsection{Discussion}

\para{Assumptions on Unlabeled Data.} In Sec. \ref{sec:formulation} we assumed that a large set $\mathcal{S}$ of $(q,c)$ pairs (without answer annotation) is readily available. We acknowledge that annotators for SQuAD dataset are shown only context $c$ and then required to provide $(q,a)$ pairs, so that $(q,c)$ pairs are not free. However, the curation of Natural Questions starts with users' information-seeking questions and draws support from information retrieval to get $(q,c)$ pairs. In this case our method has its practical value. We consider SQuAD as a testbed for our approach, while NQ fits the assumptions better. 

\para{Design efforts and time cost.} Our approach highlights efficiency during annotation, while the efforts in designing are not taken into account. We agree these efforts are non-trivial, yet they're hard to quantify. We're optimistic about efficiency since these efforts will be amortized when our approach is reused or extended to other datasets/tasks. In our study, we started with building lexicons and modules for SQuAD, but we didn't make additional efforts when we adapted to NQ. This demonstrates flexibility across different datasets. To extend our work to new tasks, some components in our study may be reused, and we hope users can learn from our experience to expedite their customization.

\para{Results with 36/54 explanations on NQ.} It is observed that on NQ dataset (Tabel \ref{tab:nq-result}), using 36 and 54 explanations both achieves 41\% F1 score. We conjecture part of the reason to be random subsampling of expalantions from a larger pool, since (1) each explanation has different representation power and generalization ability; (2) subsampled explanations could describe similar things and lack diversity. Our discussion on matching quality/bias (Sec. \ref{sec:bias}) may also account for this. We think ensuring diversity during explanation collection and enforcing instance weighting during training may help alleviate these issues, but will leave this as future work.

\section{Related Work}
\noindent
\textbf{Learning with Explanations.} \citet{srivastava2017joint} first propose to use explanations as \textit{feature} functions in concept learning. \citet{hancock2018training} proposed \textsc{BabbleLabble} for training classifiers with explanations in data programming setting, which uses explanations to provide \textit{labels} instead of features. \citet{wang2019learning} proposed \textsc{NExT} to improve generalization of explanations with softened rule execution. Both \textsc{BabbleLabble} and \textsc{NExT} highlight annotation efficiency in low-resource settings. To the best of our knowledge, we are the first to study soliciting explanations for MRC, which is intrinsically more challenging than classification tasks in existing works. Concurrent with our work, \citet{lamm2020qed} proposed \textsc{QED}, a linguistically-grounded framework for QA explanations, which decomposes the steps to answer questions into discrete steps associated with linguistic phenomena. Related to our work, \citet{dua-etal-2020-benefits} collect context spans that ``should be aggregated to answer a question'' and use these annotations as auxiliary supervision. 

\paragraph{Learning from Unlabeled data.} 
A notable line of work focuses on enforcing consistency on unlabeled data by regularizing model predictions to be invariant to noise-augmented data \cite{xie2019unsupervised,yu2018qanet}. Consistency can also be enforced through temporal ensemble \cite{laine2016temporal, tarvainen2017mean}. Another line of work uses bootstrapping -- first training a weak model with labeled data; then use model prediction on unlabeled data as supervision \cite{Carlson2009LearningAN,yang2018distantly}. Our proposed method is non-conflicting with semi-supervised strategies and we enhance NMTeacher with these strategies to achieve the best performance.

\paragraph{Neural Module Networks.} 
Neural module networks (NMNs) are dynamically composed of individual modules of different capabilities. It was first proposed for VQA tasks  \cite{andreas2016neural, andreas2016learning, hu2017learning}. Recently in NLP community, reading comprehension requiring reasoning \cite{yang2018hotpotqa, Dua2019DROP, amini2019mathqa} are proposed and widely studied. Recent works \cite{jiang2019self, gupta2020neural} generally adopt a parser that gives a sequence of operations to derive the final answer. Our work differs in that (1) operations are constructed from explanations instead of questions; (2) NMTeacher provides supervision, instead of being used as final MRC model and trained in a fully-supervised manner. We limit our scope to SQuAD-style MRC tasks in this paper and leave other challenging tasks as future work.

\paragraph{Unsupervised and Few-shot Learning for MRC.} Several lines of work share the same goal of reducing dependency on human annotation for MRC. This goal can be approached from different perspectives. (1) ``Distant'' Supervision: to generate ``proxy'' training examples automatically \cite{dhingra-etal-2018-simple, lewis-etal-2019-unsupervised, li-etal-2020-harvesting}; (2) Learning Efficiency: a model learns quickly with minimal supervision \cite{Radford2019LanguageMA, chan2019kermit}; (3) Annotation Efficiency: to create a dataset efficiently with time limit/budget; our work falls into this category. We believe these perspectives are non-conflicting with each other. It would be interesting to see whether and how methods from these perspectives can be integrated, and we leave this as future work.

\section{Conclusion}
In this paper, we propose to teach extractive MRC with explanations, with a focus on annotation efficiency.
We believe explanations stating “why” and justifying “deduction process” opens up a new way to communicate human’s generalization abilities to MRC model training.
We begin with a small set of semi-structured explanations and compose NMTeachers to augment training data. NMTeachers are modularized functions where each module has a strict and softened form, enabling broader coverage from each explanation. Extensive experiments on different datasets and MRC models demonstrate the efficiency of our system. 
Having achieved encouraging results for MRC, we look forward to extending this framework to tasks such as non-fact-based QA and multi-hop reasoning.

\section*{Acknowledgments}
This research is based upon work supported in part by the Office of the Director of National Intelligence (ODNI), Intelligence Advanced Research Projects Activity (IARPA), via Contract No. 2019-19051600007, United States Office Of Naval Research under Contract No. N660011924033, and NSF SMA 18-29268. The views and conclusions contained herein are those of the authors and should not be interpreted as necessarily representing the official policies, either expressed or implied, of ODNI, IARPA, or the U.S. Government. We would like to thank anonymous reviewers and collaborators in USC INK research lab for their constructive feedback.

\bibliographystyle{acl_natbib}
\bibliography{anthology,emnlp2020}

\clearpage
\appendix
\section{Case Study}
\paragraph{Strictly-matched instances.} Table \ref{tab:hard-match} shows two examples of strictly-matched instances. In the first example, the explanation specified how to answer questions similar to ``In what year did X (\texttt{sth.}) begin''. Intuitively, the answer should be a year number right after  ``since'', and the entity before ``begin'' should be a keyword. In the second example, questions following the pattern  ``when was X (\texttt{sth.}) Y (\texttt{done})'' are explained and the answer is typically a date after  ``on''. Also, the verb  ``\texttt{done}'' should be directly before  ``on'' and the answer.
\paragraph{Softly-match Instances.}
Table \ref{tab:soft-match} shows two examples of softly-matched instances. In the first example, the distance between Y and Z is three in the question, while the explanation specifies there should be less than two words between them. With \textsc{Compare} module, the correct answer is found with high confidence of $97.22\%$. In the second example, the explanation specifies Y to be an adjective phrase. With \textsc{Fill} module, a verb in the past tense, ``purified'', is also listed as a potential fit for variable Y, and this gives the correct answer ``a secret lake'' with a confidence score of $72.48\%$.
\begin{table}[h]
    \centering
    \scalebox{0.7}{
    \begin{tabular}{p{10cm}}
    \toprule
    \makecell[l{p{10cm}}]{\textbf{\underline{Reference Instance}}\\\textbf{Q:} In what year did \textcolor{Blue}{\textit{\textbf{\uline{Film Fest New Haven}}}} begin?\\\textbf{C:} ... The \textcolor{Blue}{\textit{\textbf{\uline{Film Fest New Haven}}}} has been held annually since 1995.\\\textbf{A:} 1995}  \\
    \makecell[l{p{10cm}}]{\textbf{\underline{Semi-structured Explanation}}\\X is ``\textcolor{Blue}{\textit{\textbf{\uline{Film Fest New Haven}}}}''. The question starts with ``In what year'', so the answer should be a year. ``begin'' is in the question. X is directly after ``did'' and directly before ``begin'' in the question. ``since'' is directly before the answer.} \\
    \makecell[l{p{10cm}}]{\textbf{\underline{Strictly-matched Instance}}\\\textbf{Q:} In what year did \textcolor{Blue}{\textit{\textbf{\uline{the Music of the Night}}}} begin?\\\textbf{C:} ... Since 1992 \textcolor{Blue}{\textit{\textbf{\uline{the Music of the Night}}}} has been performed in the Royal Citadel by the 29 Commando Regiment and local performers to raise money for local and military charities. ...\\\textbf{A:} 1992} \\
    \midrule
    \makecell[l{p{10cm}}]{\textbf{\underline{Reference Instance}}\\\textbf{Q:} When was Queen Victoria's \textcolor{Blue}{\uline{\textbf{\textit{funeral}}}} \textcolor{BrickRed}{\uwave{\textbf{held}}}?\\\textbf{C:} Her \textcolor{Blue}{\uline{\textbf{\textit{funeral}}}} was \textcolor{BrickRed}{\uwave{\textbf{held}}} on Saturday, 2 February, in St George's Chapel, Windsor Castle, and after two days of lying-in-state ...\\\textbf{A:} Saturday, 2 February}  \\
    \makecell[l{p{10cm}}]{\textbf{\underline{Semi-structured Explanation}}\\X is ``\textcolor{Blue}{\uline{\textbf{\textit{funeral}}}}''. Y is ``\textcolor{BrickRed}{\uwave{\textbf{held}}}''. In the question X is within 4 words after ``when was'' and Y is directly after X. ``on'' is directly before the answer. Y is within 2 words before the answer. X is \textbf{within 3 words}} left of Y. The question starts with ``when'', so the answer should be a date. \\
    \makecell[l{p{10cm}}]{\textbf{\underline{Strictly-matched Instance}}\\\textbf{Q:} When was \textcolor{Blue}{\uline{\textit{\textbf{independence}}}} \textcolor{BrickRed}{\uwave{\textbf{declared}}}?\\\textbf{C:} ... \textcolor{Blue}{\uline{\textit{\textbf{Independence}}}} was \textcolor{BrickRed}{\uwave{\textbf{declared}}} on 24 September 1973.\\\textbf{A:} 24 September 1973} \\
    \bottomrule
    \end{tabular}
    }
    \vspace{-0.1cm}
    \caption{\textbf{Examples of strictly-matched instances.}}
    \label{tab:hard-match}
    \vspace{-0.5cm}
\end{table}

\begin{table}[h]
    \centering
    \scalebox{0.7}{
    \begin{tabular}{l}
    \toprule
        \makecell[l{p{10cm}}]{\textbf{\underline{Reference Instance}}\\\textbf{Q:} Who did \textcolor{Blue}{\textit{\textbf{\uline{Estonia}}}} \textcolor{BrickRed}{\textbf{\uwave{rebel against}}} in \textcolor{PineGreen}{\doubleunderline{\textbf{1343}}}? \\\textbf{C:} ...  In \textcolor{PineGreen}{\doubleunderline{\textbf{1343}}}, the people of northern \textcolor{Blue}{\textit{\textbf{\uline{Estonia}}}} and Saaremaa \textcolor{BrickRed}{\textbf{\uwave{rebel against}}} German rule in the St. George's Night Uprising , which was put down by 1345. ...\\\textbf{A:} German rule}  \\
        \makecell[l{p{10cm}}]{\textbf{\underline{Semi-structured Explanation}}\\X is ``\textcolor{Blue}{\textit{\textbf{\uline{Estonia}}}}''. Y is ``\textcolor{BrickRed}{\textbf{\uwave{rebel against}}}''. Z is ``1343''. In the question, Y is directly after X and Z is within 2 words after Y. Z is a year. The answer directly follows Y. X is within 3 words before Y.} \\
        \makecell[l{p{10cm}}]{\textbf{\underline{Softly-matched Instance}} \\\textbf{Q:} \textcolor{Blue}{\textit{\textbf{\uline{The Slavs}}}} \textcolor{BrickRed}{\textbf{\uwave{appeared on}}} whose borders around \textcolor{PineGreen}{\doubleunderline{\textbf{the 6th century}}}?\\\textbf{C:} ... Around \textcolor{PineGreen}{\doubleunderline{\textbf{the 6th century}}}, \textcolor{Blue}{\textit{\textbf{\uline{Slavs}}}} \textcolor{BrickRed}{\textbf{\uwave{appeared on}}} Byzantine borders in great numbers. ...\\\textbf{A:} Byzantine borders (Confidence $z$ = 97.22\%)} \\
        \makecell[l{p{10cm}}]{\textbf{\underline{Note}}\\Z (the 6th century) is 3 words after Y (appeared on) in the question, which slightly breaks the constraint ``Z is within 2 words after Y''. This is captured by \textsc{Compare} module.}\\
        \midrule
        \makecell[l{p{10cm}}]{\textbf{\underline{Reference Instance}}\\Q: Where is \textcolor{Blue}{\textit{\textbf{\uline{hydrogen}}}} \textcolor{BrickRed}{\textbf{\uwave{highly soluble}}}?\\C: ... \textcolor{Blue}{\textit{\textbf{\uline{Hydrogen}}}} is \textcolor{BrickRed}{\textbf{\uwave{highly soluble}}} in many rare earth and transition metals and is soluble in both nanocrystalline and amorphous metals. ... \\ A: many rare earth and transition metals}\\
        \makecell[l{p{10cm}}]{\textbf{\underline{Semi-structured Explanation}}\\X is  ``\textcolor{Blue}{\textit{\textbf{\uline{hydrogen}}}}''. Y is  ``\textcolor{BrickRed}{\textbf{\uwave{highly soluble}}}''. Y is directly after X and X is directly after ``where is'' in the question. X is within 5 words before Y. Y is within 2 words before the answer. ``in'' directly before the answer. ``is'' is between X and Y.} \\
        \makecell[l{p{10cm}}]{\textbf{\underline{Softly-matched Instance}}\\Q: Where is \textcolor{Blue}{\textit{\textbf{\uline{the divinity herself}}}} \textcolor{BrickRed}{\textbf{\uwave{purified}}}? \\C: ... Afterwards the car, the vestments, and, if you like to believe it, \textcolor{Blue}{\textit{\textbf{\uline{the divinity herself}}}}, are \textcolor{BrickRed}{\textbf{\uwave{purified}}} in a secret lake. ...\\A: a secret lake (Confidence $z$ = 72.48\%)} \\
        \makecell[l{p{10cm}}]{\textbf{\underline{Note}}\\In the reference instance, Y (highly soluble) is supposed to be an adjective phrase. In the new instance, \textsc{Fill} module suggested  ``purified'' to be a promising candidate for variable Y.}\\
    \bottomrule
    \end{tabular}
    }
    \vspace{-0.1cm}
    \caption{\textbf{Examples of softly-matched instances.}}
    \label{tab:soft-match}
\end{table}

\section{Additional Performance Analysis}
\label{sec:additional_exp}
\paragraph{Performance of Fill and Find module}
\begin{table}[t]
    \centering
    \scalebox{0.85}{
    \begin{tabular}{c|c c c c}
         \toprule
         \textbf{Recall@n (\%)} & Top-$1$ & Top-$3$ & Top-$5$ & Top-$10$ \\ 
         \midrule
         \multicolumn{1}{l|}{Fill (Questions)} & 68.50 & 88.01 & 94.66 & 98.93 \\ 
         \multicolumn{1}{l|}{Fill (Contexts)} & 95.64 & 97.45 & 98.22 & 98.73 \\
         \multicolumn{1}{l|}{Find} & 41.00 & - & - & - \\
         \bottomrule
    \end{tabular}
    }
    \caption{Evaluation on Fill and Find module. We evaluate Fill on the test split (described in Sec.~\ref{sec:fill_find_pretrain}) and Find on collected explanations and their reference instances.}
    \label{tab:fill_find_results}
\end{table}

The \textsc{Fill} module is evaluated on the test split of hard-matched question pairs and context pairs, as described in Sec.~\ref{sec:fill_find_pretrain}. The \textsc{Find} module is evaluated through manual inspection on model's predictions on instances in $\mathcal{S}_o$.
For each sentence in the test set, we enumerate all possible constituents, let the model rank these spans. We take top-$n$ ($n = 1, 3, 5, 10$ for \textsc{Fill} module and $n=1$ for \textsc{Find} module) spans as output. We use recall (at $n$) $r_n=\frac{p}{q}$ as metric for evaluation, where $p$ is the number of correct spans found in top-$n$ outputs and $q$ is the number of all correct spans. Evaluation results for \text{Fill} and \text{Find} module are shown in Table~\ref{tab:fill_find_results}. As $n$ gets large, the top-$n$ outputs from the \textsc{Fill} module are able to identify most of the correct spans.


\paragraph{Further Analysis on Matching Quality.} To examine the distribution of matched data, we list the ``question heads'' in $\mathcal{S}_a$ and found the top 8 to be: when did (22.08\%); what year (8.51\%); how many (8.1\%); who was (7.27\%); what did (6.43\%); what percentage (5.39\%); what does (5.26\%); how long (4.35\%). This observation demonstrates the explanations we collect cover a wide range of question types. Yet, the distribution of input data has far more aspects than question heads. Our current implementation and design may not explain complex questions that require multi-step reasoning abilities, and this may result in strong biases in $\mathcal{S}_a$ and $\mathcal{S}_p$. 

To examine the labeling accuracy, we directly evaluate annotations obtained with the neural module teacher $\mathbb{G}$ against human annotations. On SQuAD with 52 explanations, $72.19\%$ EM and $83.35\%$ F1 is achieved on the 766 strictly-matches instances in $\mathcal{S}_a$. Noises in annotations generated with neural module teachers $\mathbb{G}$ will also cause performance downgrade in the final model $\mathbb{F}$; and thus denoising matched instances will help improve performance. Joint training may partially resolve this by encouraging consensus between $\mathbb{G}$ and MRC model $\mathbb{F}$; meanwhile we encourage future research in this direction.

\section{Beam Search Algorithm for Neural Module Teacher}

In Sec.~\ref{sec:get_answer} we mentioned the usage of beam search algorithm to search for the best combination of variable assignments. We provide the details in the Algorithm~\ref{alg:beam2}.

{
\let\oldReturn\Return
\renewcommand{\algorithmicrequire}{\textbf{Input:}}
\renewcommand{\algorithmicensure}{\textbf{Output:}}
\begin{algorithm}[t]
\begin{small}
\begin{algorithmic}[1]
\caption{Beam Search for NMTeacher}\label{alg:beam2}
   \STATE {\bfseries Input:} Neural Module Teacher $\mathbb{G}_i$, Instance $(q, c)$, Variable Candidates, Beam Width $w$, Threshold $t$
  \STATE $m = $ number of variables in $\mathbb{G}_i$
   \STATE Initialize \textsc{PrevStates}.
   \FOR{$j=1$ {\bfseries to} $m$}
   \STATE \textsc{CurrentStates} $\leftarrow\varnothing$ 
   \FOR{\textsc{State} in \textsc{PrevStates}}
   \STATE $V \leftarrow$ next unfilled variable
   \FOR{\textsc{Candidate} in (\textsc{Candidates} for \textsc{V})}
   \STATE Fill \textsc{V} in \textsc{State}
   \STATE $z\leftarrow$ confidence score of\\\hspace{1.5cm} evaluating \textsc{State} with $\mathbb{G}_i$
   \IF{$z > t$}
    \STATE \textsc{CurrStates}.append(\textsc{State})
    \ENDIF
    \ENDFOR
    \ENDFOR
    \STATE Sort (descending) \textsc{CurrStates} by $z$
    \STATE \textsc{PrevStates} $\leftarrow$ top $w$ states in \textsc{CurrStates}
    \ENDFOR
    \STATE return \textsc{CurrStates}
\end{algorithmic}
\end{small}
\end{algorithm}
}

\begin{figure*}[h]
    \centering
    \includegraphics[width=0.8\textwidth,clip,trim={0 10cm 23cm 0}]{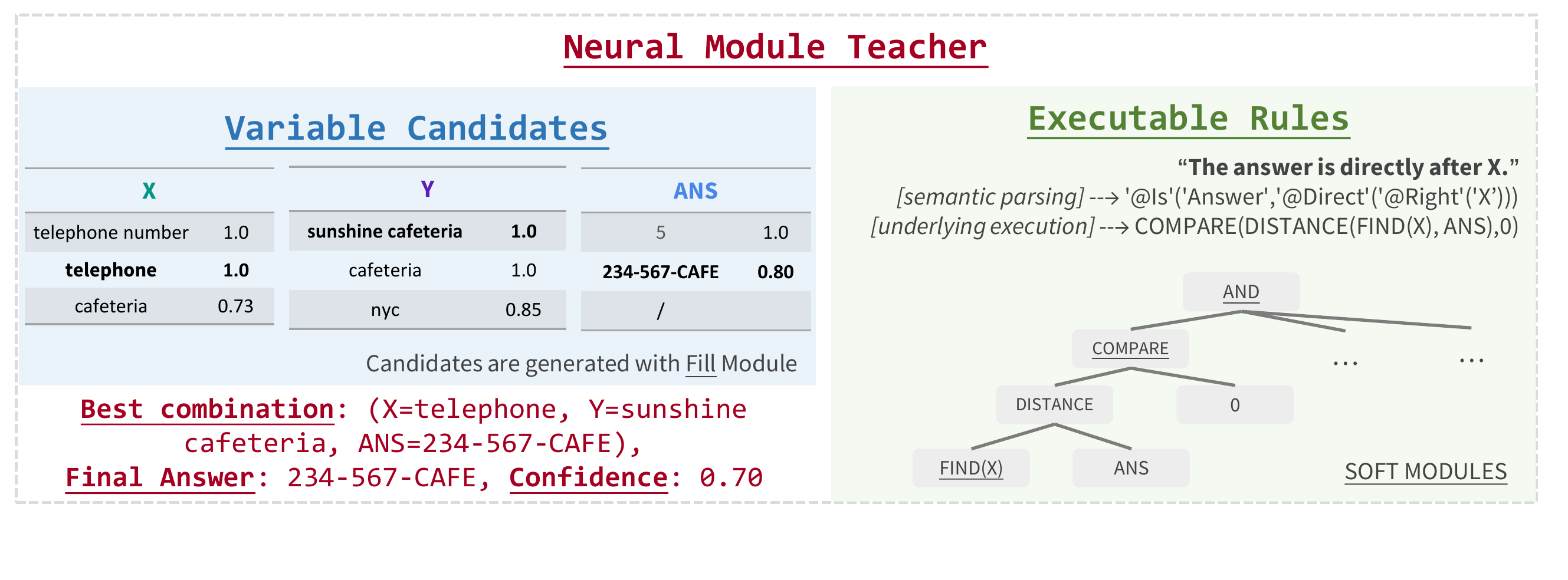}
    \vspace{-0.5cm}
    \caption{\textbf{Example for Beam Search and Extracting an Answer.} Candidates are proposed by Fill module. The best combination is selected by ranking and conducting beam search on possible combinations. Ranking is done by softened execution of rules.}
    \label{fig:beam}
\end{figure*}

\section{Reproducibility}
\paragraph{Computing Infrastructure.} Based on GPU availability, we train our models on either Quadro RT 6000, GeForce RTX 2080 Ti or GeForce GTX 1080 Ti. All of our models are trained on single GPU. NMTeacher-Joint requires optimizing both NMTeacher modules and MRC models, so we use Quadro RT 6000 for related experiments. 
\paragraph{Number of Parameters.} The two trainable modules (\textsc{Fill} and \textsc{Find}) adopt BERT-base as backbone, using 110 million parameters for each. We use several downstream MRC models in our experiments, and BERT-large is the biggest among all (340 million). To sum, NMTeacher-Joint uses 560 million parameters at most.
\paragraph{Hyper-parameters.} We use Adam with linear warmup as our optimizer and we tuned learning rate in the range of $\{1e-5, 2e-5, 3e-5, 4e-5, 5e-5\}$. We set the warmup steps to be either 100 or 500. We tuned the loss balancing co-efficient $\beta$ (in $\mathcal{L}_{ap}$ and $\mathcal{L}_{au}$, see Sec.~\ref{sec:downstream}) in the range of $\{0.1, 0.2, 0.3, 0.4, 0.5\}$. We adopt a greedy tuning strategy: first select the best learning rate and fix it; then select the best co-efficient $\beta$. We select parameters based on F1 score on dev set. 

We set the rotation interval $r$ (see Sec.~\ref{sec:downstream}) to be 8. We use batch size of $12$ for BERT-l; $16$ for BERT-b; $16$ for ALBERT-b. Gradient accumulation is adopted to achieve such batch size with GPU memory constraint.

\paragraph{Datasets.} We download both datasets we use from official websites. SQuAD: \url{https://rajpurkar.github.io/SQuAD-explorer/}; Natural Questions: \url{https://ai.google.com/research/NaturalQuestions/download}. Note that we customized the settings of NQ dataset as we limit our scope to MRC task. We aim to analysis the capability of NMTeacher in different scenarios, and thus we choose not to use the official test set due to submission constraints (\textit{e.g.}, one attempt per week). We create our own dev and test set (see Sec.~\ref{sec:exp_setup}).

\paragraph{Development Set Performance.} Table~\ref{tab:squad-result} and \ref{tab:nq-result} in the main paper lists test set performance, while their corresponding development set performance can be found in Table~\ref{tab:squad-result-dev} and \ref{tab:nq-result-dev}.

\section{Explanation Collection}
\label{sec:exp_collect_detail}
\begin{figure*}
    \centering
    \includegraphics[width=\textwidth,clip,trim={1cm, 3cm, 1cm, 2cm}]{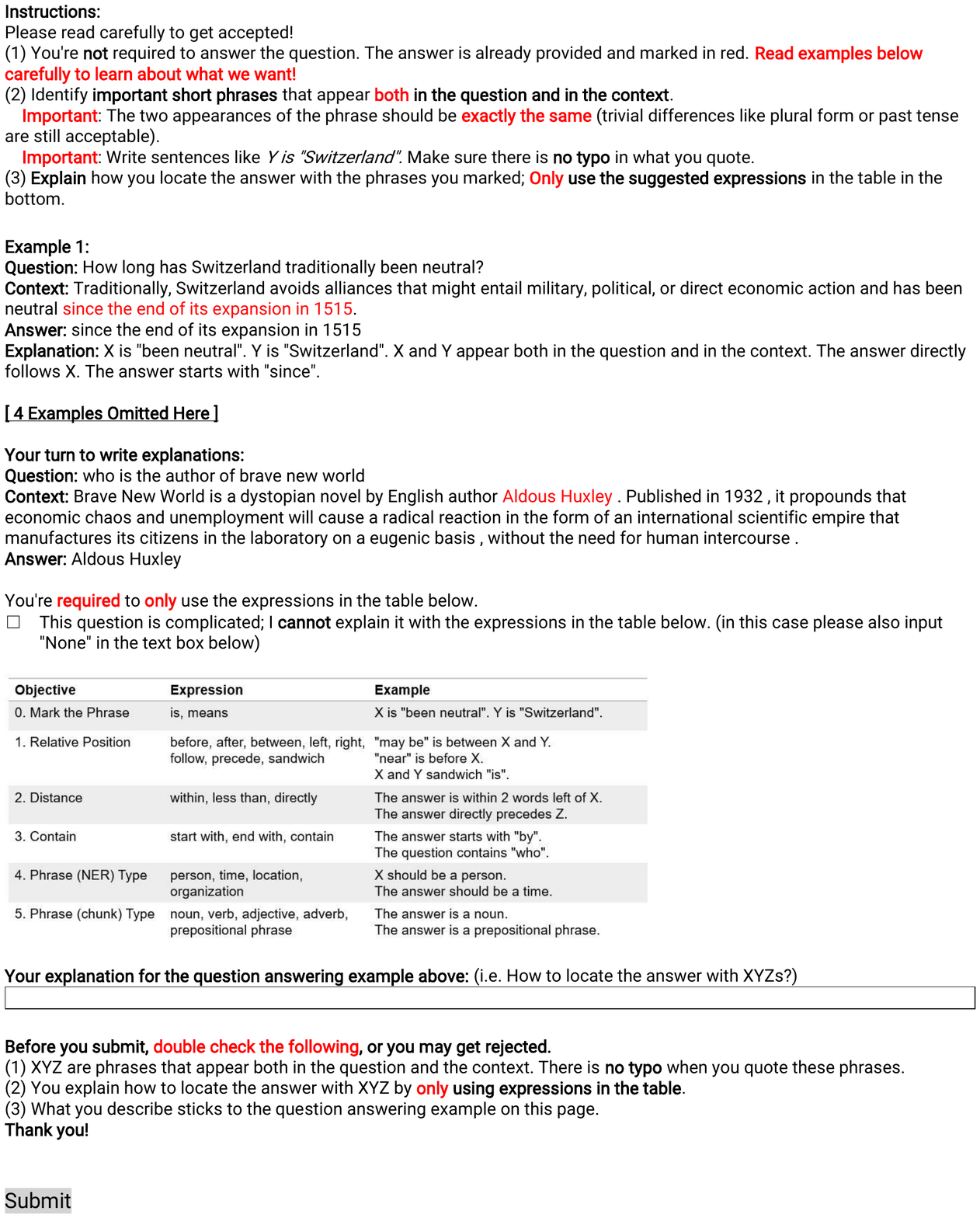}
    \caption{\textbf{Crowd-sourcing Interface on Amazon Mechanical Turk.} The interface has four parts: (1) High-level instruction; (2) 5 examples; (3) QA instance requiring explanation and an input box; (4) Final check instructions.}
    \label{fig:interface}
\end{figure*}

Our interface for collecting semi-structured explanations with Amazon Mechanical Turk is shown in Figure \ref{fig:interface}. Annotators are required to first read a short paragraph of high-level instructions and then read five provided examples. After that, they are required to write an explanation for a provided answered $(q,c,a)$ triple in one single text input box, using suggested expressions in a provided table. Finally, annotators are required to double-check their explanation before they submit. The reward for each accepted explanation is $\$0.5$.

We automatically rejected responses not following instructions (e.g., not mentioning any variables, quoted words do not appear in context). Statistics of the collected explanations on SQuAD and NQ datasets are previously shown in Table~\ref{tab:exp_stats}. We constructed and modified our parser simultaneously with the explanation collection process. The accuracy of semantic parsing is $91.93\%$ by manual inspection on 35 parsed explanations (161 sentences).

\begin{table*}[tb!]
\centering
\scalebox{0.7}{
\begin{tabular}{@{}lcccccc@{}}
\cmidrule[1pt](){1-7}
\multirow{2}{*}{\#Explanations ($|\mathcal{S}_a|$, $|\mathcal{S}_p|$)} & \multicolumn{2}{c}{13 (131, 314)} & \multicolumn{2}{c}{26 (424, 1048)} & \multicolumn{2}{c}{52 (766, 2329)} \\ \cmidrule(lr){2-3} \cmidrule(lr){4-5} \cmidrule(lr){6-7}
& EM & F1 & EM & F1 & EM & F1 \\ 
\cmidrule{1-7}
BiDAF ($\mathcal{S}_a$)  &   3.68 $\pm$ 0.82 & 7.40 $\pm$ 0.61 & 4.68 $\pm$ 0.57 & 9.39 $\pm$ 0.22 & 8.31 $\pm$ 0.55 & 13.99 $\pm$ 1.01 \\
~~+ NMTeacher-DA ($\mathcal{S}_p$)      & 4.89 $\pm$ 0.18 & 8.31 $\pm$ 0.12 & 6.24 $\pm$ 0.07 & 11.29 $\pm$ 0.20 & 13.58 $\pm$ 1.51 & 21.80 $\pm$ 2.15 \\
\cmidrule{1-7}
BERT-b ($\mathcal{S}_a$)       & 11.70 $\pm$ 0.88 & 19.11 $\pm$ 1.28 & 22.32 $\pm$ 0.24 & 33.11 $\pm$ 0.47 & 32.22 $\pm$ 1.81 & 42.68 $\pm$ 2.58  \\
~~+ NMTeacher-DA ($\mathcal{S}_p$)  & 15.68 $\pm$ 1.10 & 25.43 $\pm$ 0.98 & 24.88 $\pm$ 3.01 & 35.65 $\pm$ 4.63 & 35.67 $\pm$ 3.23 & 46.86 $\pm$ 3.41 \\
\cmidrule{1-7}
BERT-l ($\mathcal{S}_a$)       & 15.51 $\pm$ 1.61 & 23.65 $\pm$ 2.69 & 29.50 $\pm$ 2.00 & 42.05 $\pm$ 2.23 & 39.03 $\pm$ 0.63 & 51.90 $\pm$ 0.52 \\
~~+ NMTeacher-DA ($\mathcal{S}_p$)  & 18.67 $\pm$ 1.94 & 30.87 $\pm$ 2.84 & 32.76 $\pm$ 2.38 & 46.52 $\pm$ 3.22 & 43.87 $\pm$ 2.36 & 56.60 $\pm$ 2.41 \\
~~+ Self Training ($\mathcal{S}_u$)      & 15.59 $\pm$ 1.48 & 23.19 $\pm$ 1.78 & 35.48 $\pm$ 7.93 & 45.07 $\pm$ 6.04 & 46.14 $\pm$ 3.30 & 57.83 $\pm$ 3.81 \\
~~+ Mean Teacher ($\mathcal{S}_u$)      & 13.28 $\pm$ 2.48 & 21.54 $\pm$ 3.15 & 35.27 $\pm$ 4.87 & 48.35 $\pm$ 4.32 & 45.75 $\pm$ 7.14 & 58.82 $\pm$ 5.65 \\
~~+ Pseudo Labeling, PL ($\mathcal{S}_u$) & 15.96 $\pm$ 2.45 & 23.51 $\pm$ 3.66 & 41.36 $\pm$ 5.59 & 53.71 $\pm$ 7.26 & 52.95 $\pm$ 2.26 & 65.10 $\pm$ 1.14 \\
~~+ NMTeacher-Joint ($\mathcal{S}_p+\mathcal{S}_u$)& 37.06 $\pm$ 13.64 & 46.83 $\pm$ 17.34 & \textbf{61.27 $\pm$ 1.93} & 73.71 $\pm$ 2.81 & 62.22 $\pm$ 0.46 & 74.22 $\pm$ 1.24 \\
\cmidrule{1-7}
ALBERT-b ($\mathcal{S}_a$)   & 32.92 $\pm$ 1.59 & 45.62 $\pm$ 1.27 & 43.65 $\pm$ 1.63 & 57.12 $\pm$ 2.82 & 48.81 $\pm$ 1.73 & 62.06 $\pm$ 0.17  \\
~~+ NMTeacher-DA ($\mathcal{S}_p$) & 37.66 $\pm$ 2.36 & 50.25 $\pm$ 1.99 & 44.97 $\pm$ 1.20 & 58.60 $\pm$ 1.02 & 51.35 $\pm$ 2.07 & 64.27 $\pm$ 0.75 \\
~~+ Self Training ($\mathcal{S}_u$)  & 37.67 $\pm$ 4.36 & 48.32 $\pm$ 4.74 & 49.88 $\pm$ 3.06 & 61.81 $\pm$ 3.54 & 52.08 $\pm$ 2.45 & 65.34 $\pm$ 2.87 \\
~~+ Mean Teacher ($\mathcal{S}_u$)   & 33.16 $\pm$ 2.95 & 45.42 $\pm$ 2.01 &  43.42 $\pm$ 2.58 & 58.14 $\pm$ 1.74 & 56.86 $\pm$ 1.75 & 70.67 $\pm$ 1.52 \\
~~+ Pseudo Labeling, PL ($\mathcal{S}_u$) & 31.02 $\pm$ 3.32 & 43.88 $\pm$ 4.76 & 42.63 $\pm$ 2.56 & 55.62 $\pm$ 2.72 & 48.28 $\pm$ 1.63 & 60.45 $\pm$ 2.45 \\
~~+ NMTeacher-Joint ($\mathcal{S}_p+\mathcal{S}_u$) & \textbf{42.40 $\pm$ 7.47} & \textbf{56.60 $\pm$ 6.57} & 60.21 $\pm$ 3.05 & \textbf{74.44 $\pm$ 2.64} & \textbf{62.48 $\pm$ 1.23} & \textbf{75.76 $\pm$ 0.77} \\
\cmidrule[1pt](){1-7}
\end{tabular}
}
\caption{\textbf{Performance on the development set on SQuAD dataset using 13/26/52 explanations.}\label{tab:squad-result-dev}}
\end{table*}

\begin{table*}[t]
\centering
\scalebox{0.7}{
\begin{tabular}{@{}lcccccc@{}}
\cmidrule[1pt](){1-7}
\multirow{2}{*}{\#Explanations ($|\mathcal{S}_a|$, $|\mathcal{S}_p|$)} & \multicolumn{2}{c}{18 (98, 539)} & \multicolumn{2}{c}{36 (107, 647)} & \multicolumn{2}{c}{54 (273, 1047)} \\ \cmidrule(lr){2-3} \cmidrule(lr){4-5} \cmidrule(lr){6-7}
& EM & F1 & EM & F1 & EM & F1 \\
\cmidrule{1-7}
BERT-l ($\mathcal{S}_a$)       & 12.33 $\pm$ 2.28 & 22.08 $\pm$ 2.55 & 15.18 $\pm$ 0.35 & 24.89 $\pm$ 1.97 & 14.62 $\pm$ 0.77 & 24.46 $\pm$ 1.02 \\
~~+ NMTeacher-DA ($\mathcal{S}_p$)  & 17.12 $\pm$ 1.04 & 28.20 $\pm$ 0.90 & 19.60 $\pm$ 1.45 & 31.05 $\pm$ 1.70 & 20.10 $\pm$ 1.13 & 31.48 $\pm$ 1.49 \\
~~+ Self Training ($\mathcal{S}_u$)  & 15.76 $\pm$ 2.07 & 25.41 $\pm$ 0.46 & 18.61 $\pm$ 1.36 & 27.77 $\pm$ 0.31 & 18.02 $\pm$ 1.04 & 26.18 $\pm$ 0.54 \\
~~+ Mean Teacher ($\mathcal{S}_u$)   & 15.68 $\pm$ 0.74 & 25.92 $\pm$ 0.59 & 17.41 $\pm$ 0.76 & 27.97 $\pm$ 1.11 & 18.64 $\pm$ 1.55 & 27.88 $\pm$ 1.69 \\
~~+ Pseudo Labeling, PL ($\mathcal{S}_u$) & 18.02 $\pm$ 2.07 & 25.64 $\pm$ 0.92 & 20.95 $\pm$ 2.52 & 28.55 $\pm$ 2.63 & 17.17 $\pm$ 0.42 & 26.12 $\pm$ 0.48 \\
~~+ NMTeacher-Joint ($\mathcal{S}_p+\mathcal{S}_u$) & 16.69 $\pm$ 0.79 & 28.48 $\pm$ 1.16 & 21.62 $\pm$ 1.82 & 32.51 $\pm$ 2.06 & 22.90 $\pm$ 2.24 & 34.02 $\pm$ 2.20 \\
\cmidrule{1-7}
ALBERT-b ($\mathcal{S}_a$)   & 20.02 $\pm$ 2.05 & 28.80 $\pm$ 2.21 & 22.90 $\pm$ 1.74 & 32.19 $\pm$ 2.22 & 21.65 $\pm$ 0.83 & 32.23 $\pm$ 1.20 \\
~~+ NMTeacher-DA ($\mathcal{S}_p$) & 21.27 $\pm$ 1.19 & 30.87 $\pm$ 1.99 & 25.80 $\pm$ 2.48 & 35.92 $\pm$ 2.78 & 23.22 $\pm$ 2.73 & 33.94 $\pm$ 2.98 \\
~~+ Self Training ($\mathcal{S}_u$)  & 19.68 $\pm$ 1.66 & 28.67 $\pm$ 2.09 & 23.64 $\pm$ 2.70 & 32.73 $\pm$ 1.79 & 23.64 $\pm$ 2.36 & 32.78 $\pm$ 2.58 \\
~~+ Mean Teacher ($\mathcal{S}_u$)   & 19.44 $\pm$ 0.12 & 28.84 $\pm$ 1.04 & 24.79 $\pm$ 2.92 & 33.96 $\pm$ 3.15 & 29.23 $\pm$ 3.63 & 38.84 $\pm$ 3.27 \\
~~+ Pseudo Labeling, PL ($\mathcal{S}_u$) & 19.04 $\pm$ 1.29 & 27.35 $\pm$ 2.11 & 22.98 $\pm$ 2.47 & 31.48 $\pm$ 1.29 & 20.34 $\pm$ 0.92 & 31.07 $\pm$ 2.59 \\
~~+ NMTeacher-Joint ($\mathcal{S}_p+\mathcal{S}_u$) & \textbf{24.44 $\pm$ 4.08} & \textbf{35.09 $\pm$ 5.30} & \textbf{31.12 $\pm$ 2.38} & \textbf{41.74 $\pm$ 3.56} & \textbf{29.13 $\pm$ 3.77} & \textbf{40.22 $\pm$ 3.98} \\
\cmidrule[1pt](){1-7}
\end{tabular}
}
\caption{\textbf{Performance on the development set on Natural Questions dataset using 18/36/54 explanations.}\label{tab:nq-result-dev}}
\end{table*}

\end{document}